\documentclass{article} 
\usepackage{iclr2016_conference,times}
\usepackage{hyperref}
\usepackage{url}
\usepackage{graphicx}
\usepackage{amsmath}
\usepackage{caption}
\usepackage{epstopdf}
\usepackage{algpseudocode}
\usepackage{algorithm}

\newcommand{\algrule}[1][.2pt]{\par\vskip.5\baselineskip\hrule height #1\par\vskip.5\baselineskip}

\newenvironment{itemize*}%
  {\begin{itemize}%
    \setlength{\itemsep}{1pt}%
    \setlength{\parskip}{1pt}}%
  {\end{itemize}}

\newenvironment{enumerate*}%
  {\begin{enumerate}%
    \setlength{\itemsep}{1pt}%
    \setlength{\parskip}{1pt}}%
  {\end{enumerate}}

\title{All you need is a good init}

\author{Dmytro Mishkin, Jiri Matas  \\
\\
Center for Machine Perception\\
 Czech Technical University in Prague\\
 Czech Republic
\texttt{\{mishkdmy,matas\}@cmp.felk.cvut.cz} \\
}

\iclrfinalcopy 

\begin{document}
\maketitle
\begin{abstract}
Layer-sequential unit-variance (LSUV) initialization --  a simple method for weight initialization for deep net learning -- is proposed. The method consists of the two steps. First, pre-initialize weights of each convolution or inner-product layer with orthonormal matrices. Second, proceed from the first to the final layer, normalizing the variance of the output of each layer to be equal to one.

Experiment with different activation functions (maxout, ReLU-family, tanh) show that the proposed initialization leads to learning of very deep nets that (i) produces networks with test accuracy better or equal to standard methods and (ii) is at least as fast as the complex schemes proposed specifically for very deep nets such as FitNets~(\cite{FitNets2014}) and Highway~(\cite{Highway2015}). 

Performance is evaluated on GoogLeNet, CaffeNet, FitNets and Residual nets and the state-of-the-art, or very close to it, is achieved on the MNIST, CIFAR-10/100 and ImageNet datasets.
\end{abstract}

\section{Introduction}
\label{intro}
Deep nets have demonstrated impressive results on a number of computer vision and natural language processing problems. 
At present, state-of-the-art results in image classification~(\cite{VGGNet2015,Googlenet2015}) and speech recognition~(\cite{VGGNetSound2015}), etc., have been achieved with very deep ($\geq 16$ layer) CNNs.
Thin deep nets are of particular interest, since they are accurate and at the same inference-time efficient~(\cite{FitNets2014}).

One of the main obstacles preventing the wide adoption of very deep nets is the absence of a general, repeatable and efficient procedure for their end-to-end training. 
For example, VGGNet~(\cite{VGGNet2015}) was optimized by a four stage procedure that started by training a network with moderate depth, adding progressively more layers.
\cite{FitNets2014} stated that deep and thin networks are very hard to train by backpropagation if deeper than five layers, especially with uniform initialization.

On the other hand, \cite{MSRA2015} showed that it is possible to train the VGGNet in a single optimization run if the network weights are initialized with a specific ReLU-aware initialization. The \cite{MSRA2015} procedure 
generalizes to the ReLU non-linearity the idea of filter-size dependent initialization, introduced for the linear case by~(\cite{Xavier10}).
Batch normalization~(\cite{BatchNorm2015}), a technique that inserts layers into the the deep net that transform the output for the batch to be zero mean unit variance, has successfully facilitated training of the twenty-two layer GoogLeNet~(\cite{Googlenet2015}).
However, batch normalization adds a 30\% computational overhead to each iteration. 

The main contribution of the paper is a proposal of a simple initialization procedure that, in connection with standard stochastic gradient descent~(SGD), leads to state-of-the-art thin and very deep neural nets\footnote{The code allowing to reproduce the experiments is available at \\ \url{https://github.com/ducha-aiki/LSUVinit}}.
The result highlights the importance of initialization in very deep nets. We review the history of CNN initialization in Section~\ref{sec:initialization-review}, which is followed by a detailed description of the novel initialization method in Section~\ref{sec:algorithm}. The method is experimentally validated in Section~\ref{sec:experiment}.

\section {Initialization in neural networks}
\label{sec:initialization-review}

After the success of CNNs in IVSRC 2012~(\cite{AlexNet2012}), initialization with Gaussian noise with mean equal to zero and standard deviation set to 0.01 and adding bias equal to one for some layers become very popular. But, as mentioned before, it is not possible to train very deep network from scratch with it~(\cite{VGGNet2015}).
The problem is caused by the activation (and/or) gradient magnitude in final layers~(\cite{MSRA2015}). If each layer, not properly initialized, scales input by $k$, the final scale would be $k^{L}$, where $L$ is a number of layers. Values of $k>1$ lead to extremely large values of output layers,  $k<1$ leads to a diminishing signal and gradient.

\begin{figure}[tb]
\centering
\includegraphics[width=0.49\linewidth]{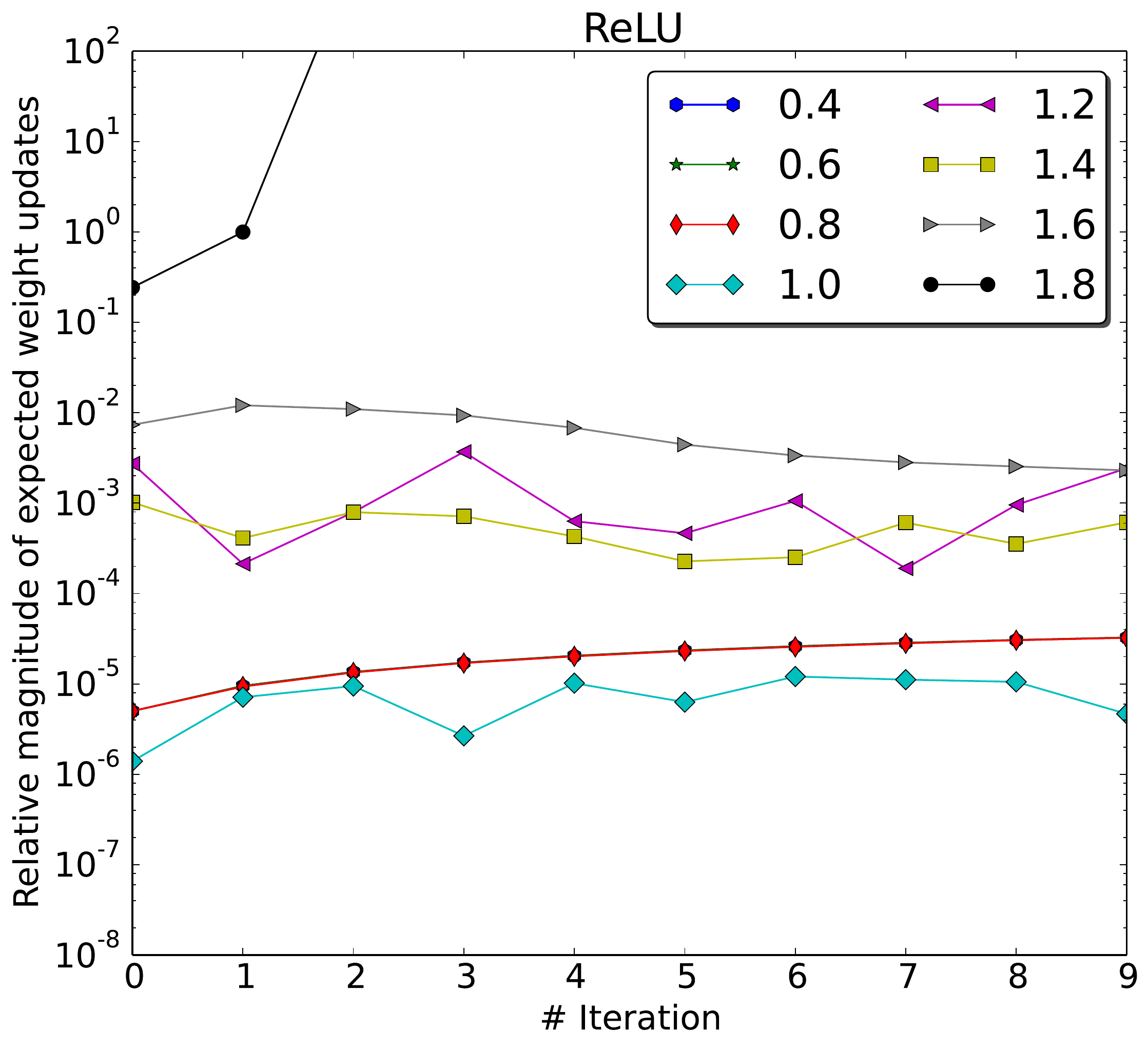}
\includegraphics[width=0.49\linewidth]{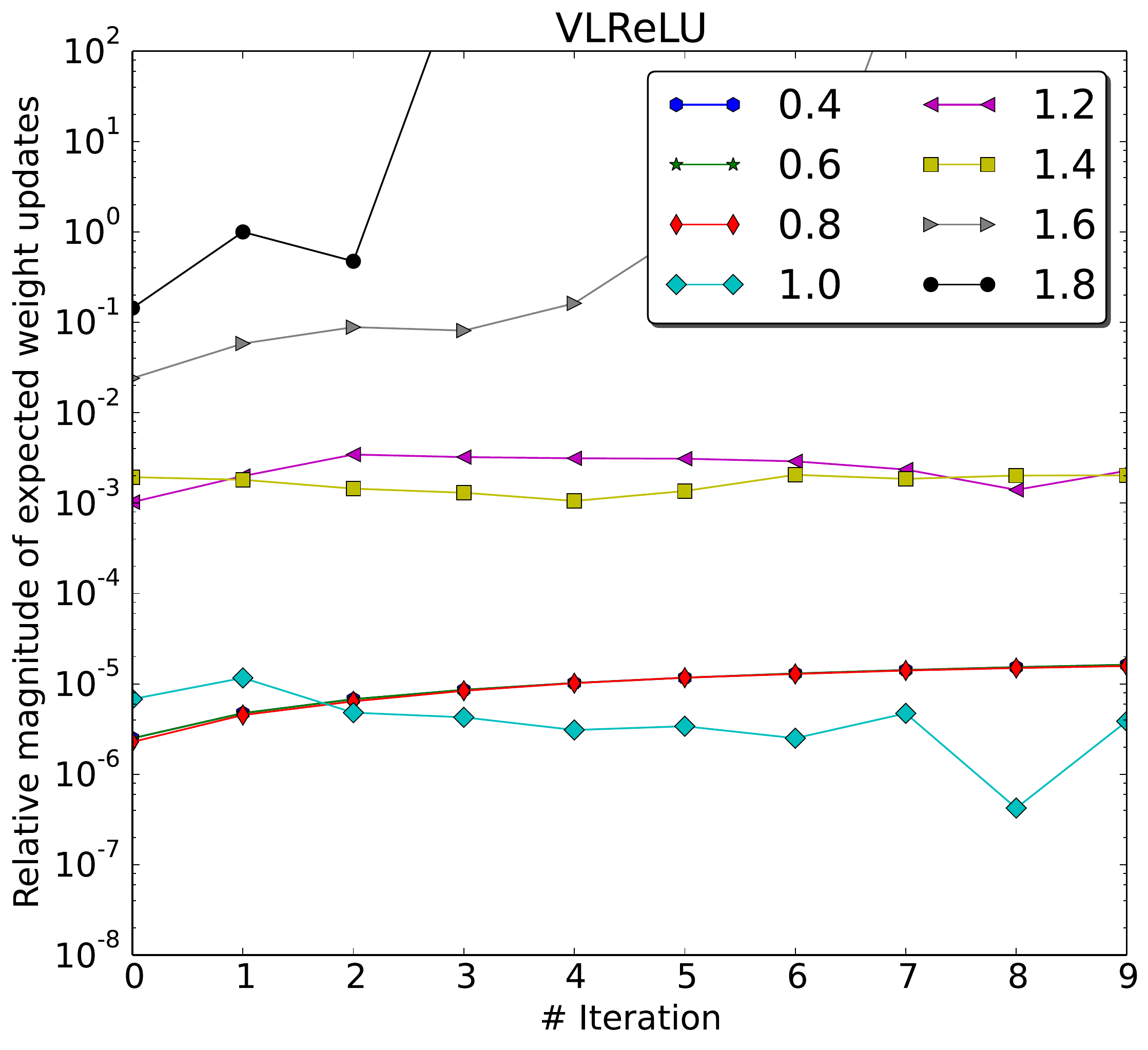}\\
\includegraphics[width=0.49\linewidth]{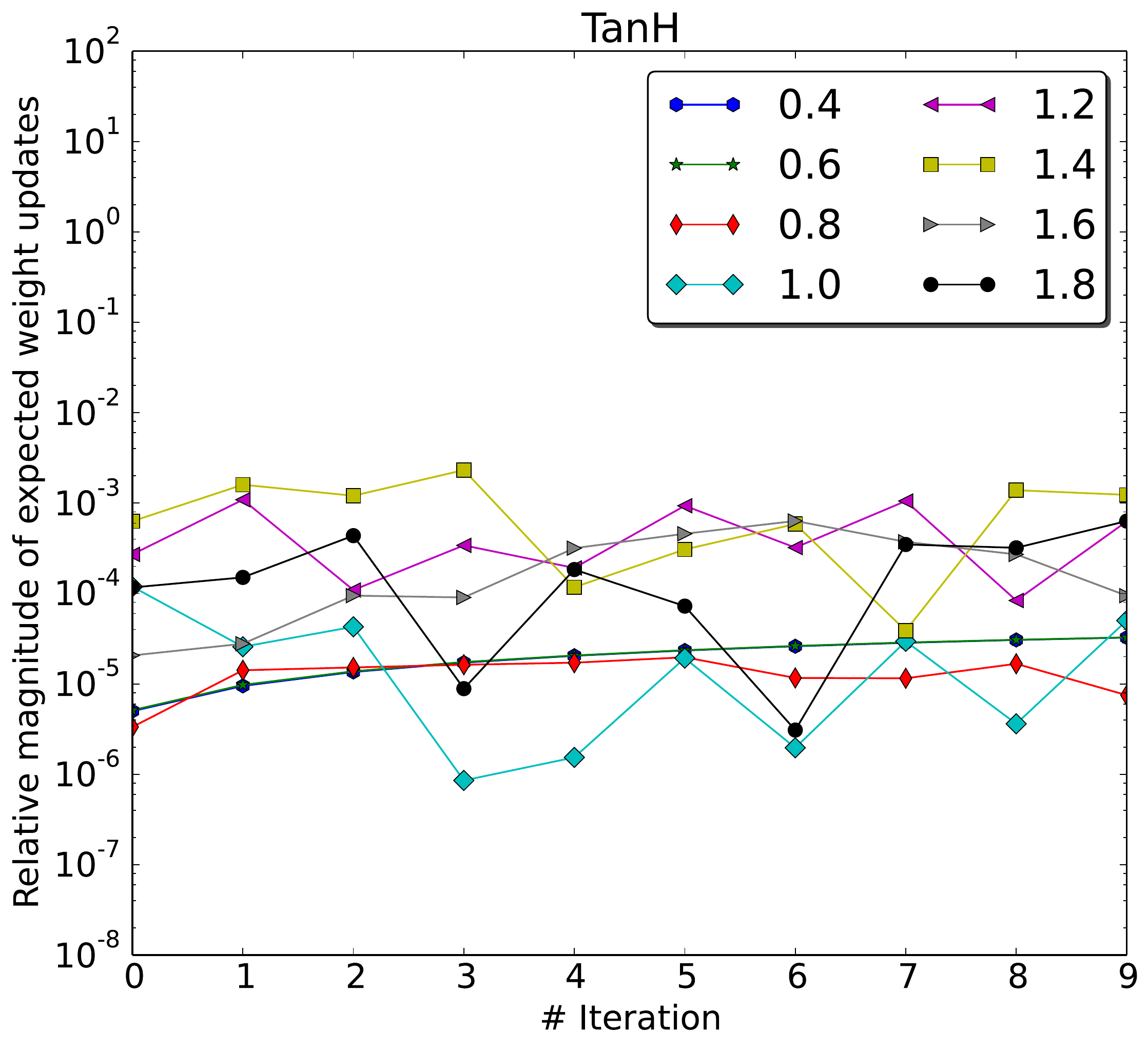}
\includegraphics[width=0.49\linewidth]{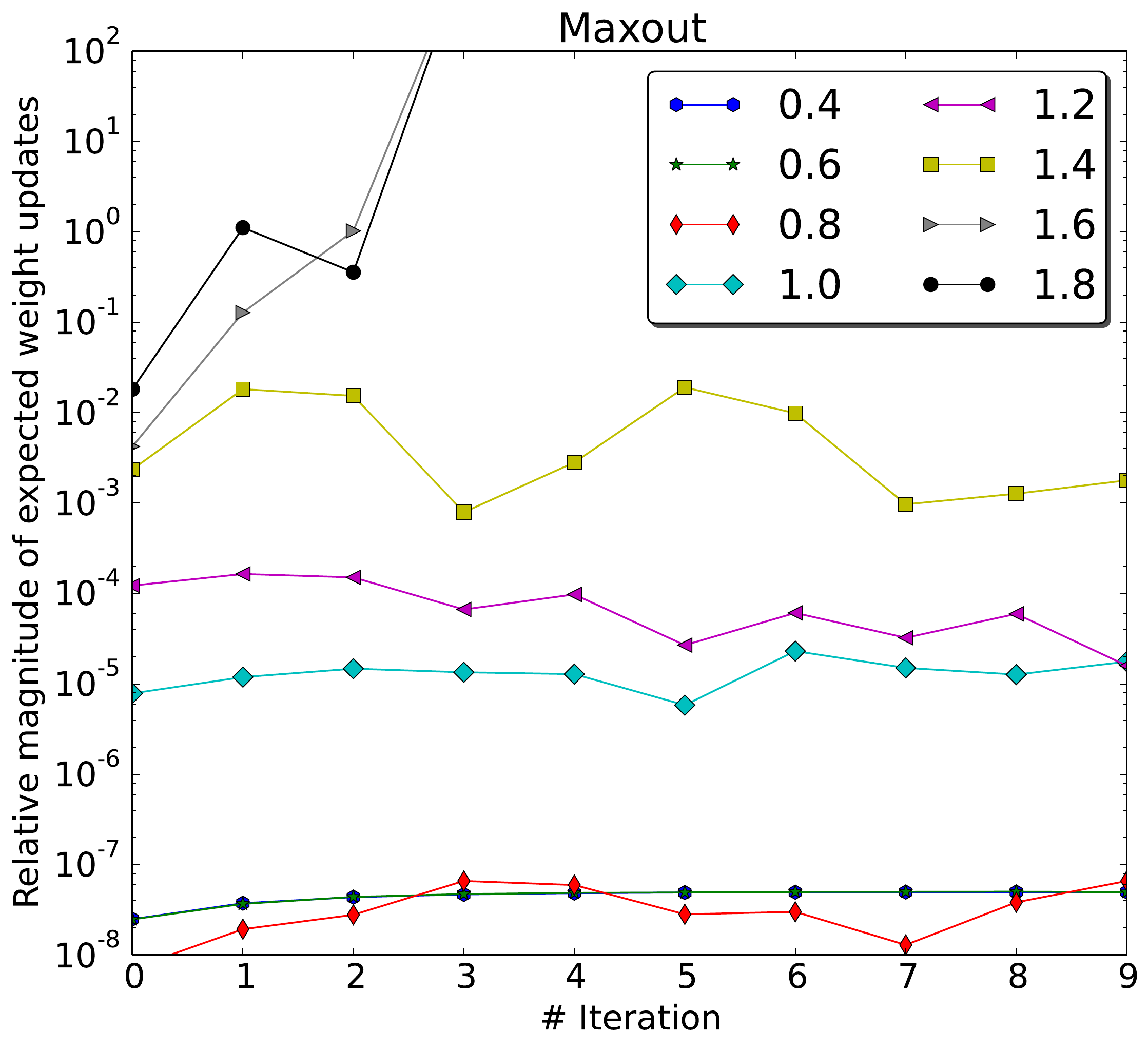}
\caption{Relative magnitude of weight updates as a function of the training iteration for different weight initialization scaling after ortho-normalization. The values in the range 0.1\% .. 1\% lead to convergence, larger to divergence, for smaller, the network  can hardly leave  the initial state. Subgraphs show results for different non-linearities -- ReLU (top left), VLReLU (top right), hyperbolic tangent (bottom left) and Maxout (bottom right).}
\label{fig:weights-scaling}
\end{figure}

\cite{Xavier10} proposed a formula for estimating the standard deviation on the basis of the number of input and output channels of the layers under assumption of no non-linearity between layers. Despite invalidity of the  assumption, Glorot initialization works well in many applications. 
~\cite{MSRA2015} extended this formula to the ReLU~(\cite{ReLU2011}) non-linearity and showed its superior performance for ReLU-based nets. 
Figure~\ref{fig:weights-scaling} shows why scaling is important. Large weights lead to divergence via updates larger than the initial values, small initial weights do not allow the network to learn since the updates are of the order of 0.0001$\%$ per iteration. The optimal scaling for ReLU-net is around 1.4, which is in line with the theoretically derived $\sqrt{2}$ by \cite{MSRA2015}. 
\cite{Sussillo2014}  proposed the so called Random walk initialization, RWI, which keeps constant the log of the norms of the backpropagated errors. In our experiments, we have not been able to obtain good results with our implementation of RWI, that is why this method is not evaluated in experimental section.

\cite{KDHinton2015} and \cite{FitNets2014} take another approach to initialization and formulate training as mimicking teacher network predictions (so called knowledge distillation) and internal representations (so called Hints initialization) rather than minimizing the softmax loss.

~\cite{Highway2015} proposed a LSTM-inspired gating scheme to control information and gradient flow through the network. They trained a 1000-layers MLP network on MNIST. Basically, this kind of networks implicitly learns the depth needed for the 
given task. 

Independently, \cite{OrthoNorm2013} showed that orthonormal matrix initialization works much better for linear networks than Gaussian noise, which is only approximate orthogonal. It also work for networks with non-linearities.

The approach of layer-wise pre-training~(\cite{Bengio2006}) which is still useful for multi-layer-perceptron, is not popular for training discriminative convolution networks. 
\section{Layer-sequential unit-variance initialization}
\label{sec:algorithm}
To the best of our knowledge, there have been no attempts to generalize~\cite{Xavier10} formulas to non-linearities other than ReLU, such as tanh, maxout, etc. Also, the formula does not cover max-pooling, local normalization layers~\cite{AlexNet2012} and other types of layers which influences activations variance.
Instead of theoretical derivation for all possible layer types, or doing extensive parameters search as in Figure~\ref{fig:weights-scaling}, we propose a data-driven weights initialization.

We thus extend the orthonormal initialization~\cite{OrthoNorm2013} to an iterative procedure, described in Algorithm~\ref{alg:LSUV}. \cite{OrthoNorm2013} could be implemented in two steps. First, fill the weights with Gaussian noise with unit variance. Second, decompose them to orthonormal basis with QR or SVD-decomposition and replace weights with one of the components. 

The LSUV process then estimates output variance of each convolution and inner product layer and scales the weight to make variance equal to one. The influence of selected mini-batch size on estimated variance is negligible in wide margins, see Appendix.

The proposed scheme can be viewed as an orthonormal initialization combined with batch normalization performed only on the first mini-batch. The similarity to batch normalization is the unit variance normalization procedure, while initial ortho-normalization of weights matrices efficiently de-correlates layer activations, which is not done in~\cite{BatchNorm2015}. Experiments show that such normalization is sufficient and computationally highly efficient in comparison with full batch normalization.

\begin{algorithm}[b]
\caption{Layer-sequential unit-variance orthogonal initialization.
$L$ -- convolution or full-connected layer, $W_L$ -  its weights, $B_L$ - its output blob.,
$\textit{Tol}_\textit{var}$ - variance tolerance,  $T_i$ -- current trial, $T_\textit{max}$ -- max number of trials.}
\label{alg:LSUV}
\begin{algorithmic}
\footnotesize
\algrule
\State {\bf Pre-initialize} network with orthonormal matrices as in ~\cite{OrthoNorm2013}
\For {each layer $L$}
\While { $|\textit{Var}(B_L)  - 1.0| \ge \textit{Tol}_\textit{var} \text{ and } (T_i < T_\textit{max})$} 
\State do Forward pass with a mini-batch
\State calculate $\textit{Var}(B_L)$ 
\State $W_L$ = $W_L$ / $\sqrt{\textit{Var}(B_L)}$
\EndWhile
\EndFor
\end{algorithmic}
\end{algorithm}

The LSUV algorithm is summarized in Algorithm~\ref{alg:LSUV}.
The single parameter $\textit{Tol}_\textit{var}$ influences convergence of the initialization procedure, not the properties of the trained network. Its value does not noticeably influence the performance in a broad range of 0.01 to 0.1. 
 Because of data variations, it is often not possible to normalize variance with the desired precision. To eliminate the possibility of an infinite loop, we restricted number of trials to $T_\textit{max}$. However, in experiments described in paper, the $T_\textit{max}$ was never reached. The desired variance was achieved in 1-5 iterations.
 
We tested a variant LSUV initialization which was normalizing input activations of the each layer instead of output ones. Normalizing the input or output is identical for standard feed-forward nets, but normalizing input is much more complicated for networks with maxout~(\cite{Maxout2013}) or for networks like GoogLeNet~(\cite{Googlenet2015}) which use the output of multiple layers as input. Input normalization brought no improvement of results when tested against the LSUV Algorithm~\ref{alg:LSUV},

LSUV was also tested with pre-initialization of weights with Gaussian noise instead of orthonormal matrices. The Gaussian initialization led to small, but consistent, decrease in performance.  
\section{Experimental validation}
\label{sec:experiment}
Here we show that very deep and thin nets could be trained in a single stage. Network architectures are exactly as proposed by~\cite{FitNets2014}. The architectures are presented in Table~\ref{tab:fitnet-architectures}.
\begin{table}[htb]
\caption{ FitNets~\cite{FitNets2014} network architecture used in experiments. Non-linearity: Maxout with 2 linear pieces in convolution layers, Maxout with 5 linear pieces in fully-connected.}
\label{tab:fitnet-architectures}
\centering
\setlength{\tabcolsep}{.3em}
\begin{tabular}{|c|c|c|c|}
\hline
FitNet-1 & FitNet-4 & FitResNet-4& FitNet-MNIST\\
250K param & 2.5M param &  2.5M param & 30K param\\
\hline
conv 3x3x16 & conv 3x3x32 & conv 3x3x32 & conv 3x3x16 \\
conv 3x3x16 & conv 3x3x32 & conv 3x3x32 $\rightarrow$sum & conv 3x3x16 \\
conv 3x3x16 & conv 3x3x32 & conv 3x3x48 & \\
&  conv 3x3x48&  conv 3x3x48 $\rightarrow$ssum &  \\
&  conv 3x3x48&  conv 3x3x48 &  \\
pool 2x2&  pool 2x2 & pool 2x2 & pool 4x4, stride2 \\
\hline
conv 3x3x32 & conv 3x3x80 & conv 3x3x80 & conv 3x3x16 \\
conv 3x3x32 & conv 3x3x80 & conv 3x3x80 $\rightarrow$sum  & conv 3x3x16 \\
conv 3x3x32 & conv 3x3x80& conv 3x3x80 &  \\
&  conv 3x3x80& conv 3x3x80$\rightarrow$sum &  \\
&  conv 3x3x80& conv 3x3x80  &  \\
pool 2x2&  pool 2x2 & pool 2x2 & pool 4x4, stride2 \\
\hline
conv 3x3x48 & conv 3x3x128 & conv 3x3x128 & conv 3x3x12 \\
conv 3x3x48 & conv 3x3x128 & conv 3x3x128 $\rightarrow$sum& conv 3x3x12 \\
conv 3x3x64 & conv 3x3x128& conv 3x3x128  & \\
&  conv 3x3x128& conv 3x3x128 $\rightarrow$sum &  \\
&  conv 3x3x128 & conv 3x3x128 & \\
pool 8x8 (global)&  pool 8x8 (global) &pool 8x8 (global) & pool 2x2 \\
\hline
fc-500&  fc-500 &fc-500&   \\
softmax-10(100) & softmax-10(100) & softmax-10(100) & softmax-10\\
\hline
\end{tabular}
\end{table}
\subsection{MNIST}
First, as a "sanity check", we performed an experiment on the MNIST dataset~(\cite{MNIST1998}). It consists of 60,000 28x28 grayscale images of handwritten digits 0 to 9.
We selected the FitNet-MNIST architecture (see Table~\ref{tab:fitnet-architectures}) of~\cite{FitNets2014} and trained it with the proposed initialization strategy, without data augmentation. Recognition results are shown in Table~\ref{tab:CIFAR-MNIST}, right block.
LSUV outperforms orthonormal initialization and both LSUV and orthonormal outperform Hints initialization~\cite{FitNets2014}. 
The error rates of the Deeply-Supervised Nets (DSN,~\cite{DSN2015}) and maxout networks~\cite{Maxout2013}, the current state-of-art, are provided for reference.

Since the widely cited DSN error rate of 0.39\%, the state-of-the-art (until recently) was obtained after replacing the softmax classifier with SVM, we do the same and also observe improved results (line FitNet-LSUV-SVM in Table~\ref{tab:CIFAR-MNIST}).

\subsection{CIFAR-10/100} 
We validated the proposed initialization LSUV strategy on the CIFAR-10/100~(\cite{CIFAR2010}) dataset. It contains 60,000 32x32 RGB images, which are divided into 10 and 100 classes, respectively. 

The FitNets are trained with the stochastic gradient descent with momentum set to 0.9, the initial learning rate set to 0.01 and reduced by a factor of 10 after the 100th, 150th and 200th epoch, finishing at 230th epoch. \cite{Highway2015} and \cite{FitNets2014} trained their networks for 500 epochs. Of course, training time is a trade-off dependent on the desired accuracy; one could train a slightly less accurate network much faster.   

Like in the MNIST experiment, LSUV and orthonormal initialized nets outperformed Hints-trained Fitnets, leading to the new state-of-art when using commonly used augmentation -- mirroring and random shifts. The gain on the fine-grained CIFAR-100 is much larger than on CIFAR-10. Also, note that FitNets with LSUV initialization outperform even much larger networks like Large-All-CNN~\cite{ALLCNN2015} and Fractional Max-pooling~\cite{FractMaxPool2014} trained with affine and color dataset augmentation on CIFAR-100. 
The results of LSUV are virtually identical to the orthonormal initialization. 
\begin{table}[htb]
\caption{Network performance comparison on the MNIST and CIFAR-10/100 datasets. Results marked '$\dagger$' were obtained with the RMSProp optimizer~\cite{Tieleman2012}.}
\label{tab:CIFAR-MNIST}
\footnotesize
\centering
\setlength{\tabcolsep}{.3em}
\begin{tabular}{lll}
\multicolumn{3}{c}{Accuracy on CIFAR-10/100, with data augmentation}\\
\hline
Network &CIFAR-10, $[\%]$ & CIFAR-100,$[\%]$ \\
\hline
Fitnet4-LSUV & \textbf{93.94} & 70.04 (\textbf{72.34}$\dagger$) \\
Fitnet4-OrthoInit & 93.78          & 70.44 (72.30$\dagger$) \\
Fitnet4-Hints     & 91.61          & 64.96 \\
Fitnet4-Highway   & 92.46          & 68.09 \\
\hline
ALL-CNN &  92.75 & 66.29\\
DSN &  92.03 & 65.43\\
NiN &  91.19 & 64.32\\
maxout &  90.62 & 65.46\\
\textit{MIN} & \textit{93.25}& \textit{71.14} \\
\hline
\multicolumn{3}{c}{Extreme data augmentation}\\
\hline
Large ALL-CNN& 95.59 & n/a\\
Fractional MP (1 test) & 95.50 & 68.55 \\
Fractional MP (12 tests)& \textbf{96.53} & \textbf{73.61}\\
\hline
\end{tabular}
\begin{tabular}{llll}
\multicolumn{4}{c}{Error on MNIST w/o data augmentation}\\
\hline
Network &  layers &  params & Error, $\%$\\
\hline
\multicolumn{4}{c}{FitNet-like networks}\\
\hline
HighWay-16 & 10 & 39K & 0.57  \\
FitNet-Hints & 6 &30K & 0.51\\
FitNet-Ortho & 6 &30K & 0.48\\
FitNet-LSUV & 6 &30K & 0.48\\
FitNet-Ortho-SVM & 6 &30K & 0.43\\
FitNet-LSUV-SVM & 6 &30K & \textbf{0.38}\\
\hline
\multicolumn{4}{c}{State-of-art-networks}\\
\hline
DSN-Softmax & 3 & 350K & 0.51  \\
DSN-SVM & 3 & 350K & 0.39  \\
HighWay-32 & 10 & 151K & 0.45  \\
maxout & 3 & 420K & 0.45  \\
\textit{MIN}~\footnotemark &\textit{9} & \textit{447K} &  \textit{0.24}  \\
\hline
\end{tabular}
\end{table}
\footnotetext{
When preparing this submission we have found recent unreviewed paper MIN~\cite{MIN2015} paper, which uses a sophisticated combination of batch normalization, maxout and network-in-network non-linearities and establishes a new state-of-art on MNIST.}
\section{Analysis of empirical results}
\label{sec:solver-times-inits}
\subsection{Initialization strategies and non-linearities}
\begin{table}[htb]
\caption{The compatibility of activation functions and initialization.
\hspace{\textwidth}
Dataset: CIFAR-10. Architecture: FitNet4, 2.5M params for maxout net, 1.2M for the rest, 17 layers. The n/c symbol stands for ``failed to converge''; n/c$\dagger$ -- after extensive trials, we managed to train a maxout-net with MSRA initialization with very small learning rate and gradient clipping, see Figure~\ref{fig:sgd-cifar10}. The experiment is marked n/c as training time was excessive and parameters non-standard.}
\label{tab:activations}
\centering
\begin{tabular}{llllll}
\hline
Init method & maxout & ReLU  & VLReLU & tanh & Sigmoid \\
\hline
LSUV           & \textbf{93.94} & \textbf{92.11} & 92.97  & 89.28 & n/c \\
OrthoNorm      & 93.78          & 91.74          & 92.40           & 89.48 & n/c \\
OrthoNorm-MSRA scaled & --       & 91.93   & \textbf{93.09}  & -- & n/c \\
Xavier         & 91.75          & 90.63          & 92.27   & \textbf{89.82} & n/c \\
MSRA           & n/c$\dagger$   & 90.91          & 92.43           & 89.54 & n/c \\
\hline
\end{tabular}
\end{table}
 For the FitNet-1 architecture, we have not experienced any difficulties training the network with any of the activation functions (ReLU, maxout, tanh), optimizers (SGD, RMSProp) or initialization (Xavier, MSRA, Ortho, LSUV), unlike the uniform initialization used in~\cite{FitNets2014}. The most probable cause is that CNNs tolerate a wide variety of mediocre initialization, only the learning time increases.
The differences in the final accuracy between the different initialization methods for the FitNet-1 architecture is rather small and are therefore not presented here. 

The FitNet-4 architecture is much more difficult to optimize and thus we focus on it in the experiments
presented in this section.

We have explored the initializations with different activation functions in very deep networks. More specifically, ReLU, hyperbolic tangent, sigmoid, maxout and the VLReLU -- very leaky ReLU~(\cite{SparseConvNet2014}) -- a variant of leaky ReLU (~\cite{Maas2013}, with a large value of the negative slope 0.333, instead of the originally proposed 0.01) which is popular in Kaggle competitions~\cite{deepsea}, \cite{GrahamCIFAR}).

\begin{figure}
\centering
\includegraphics[width=0.48\linewidth]{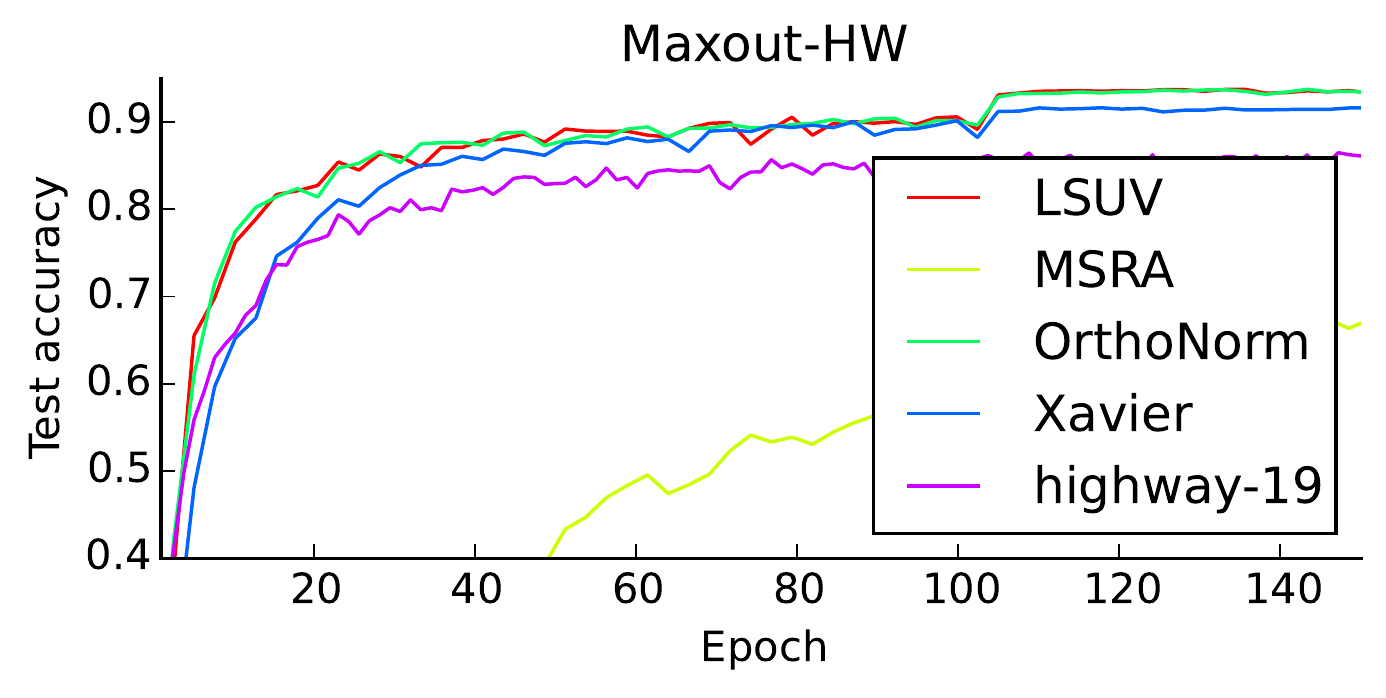}
\includegraphics[width=0.48\linewidth]{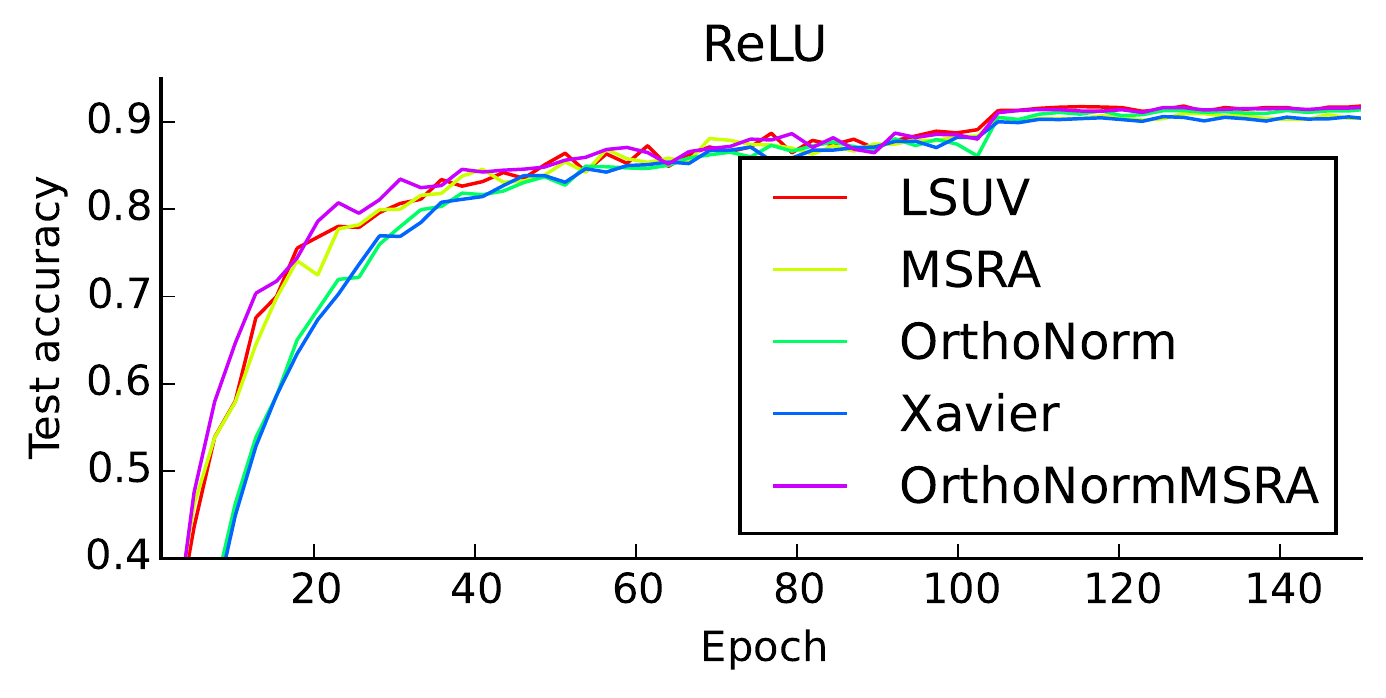}\\
\includegraphics[width=0.48\linewidth]{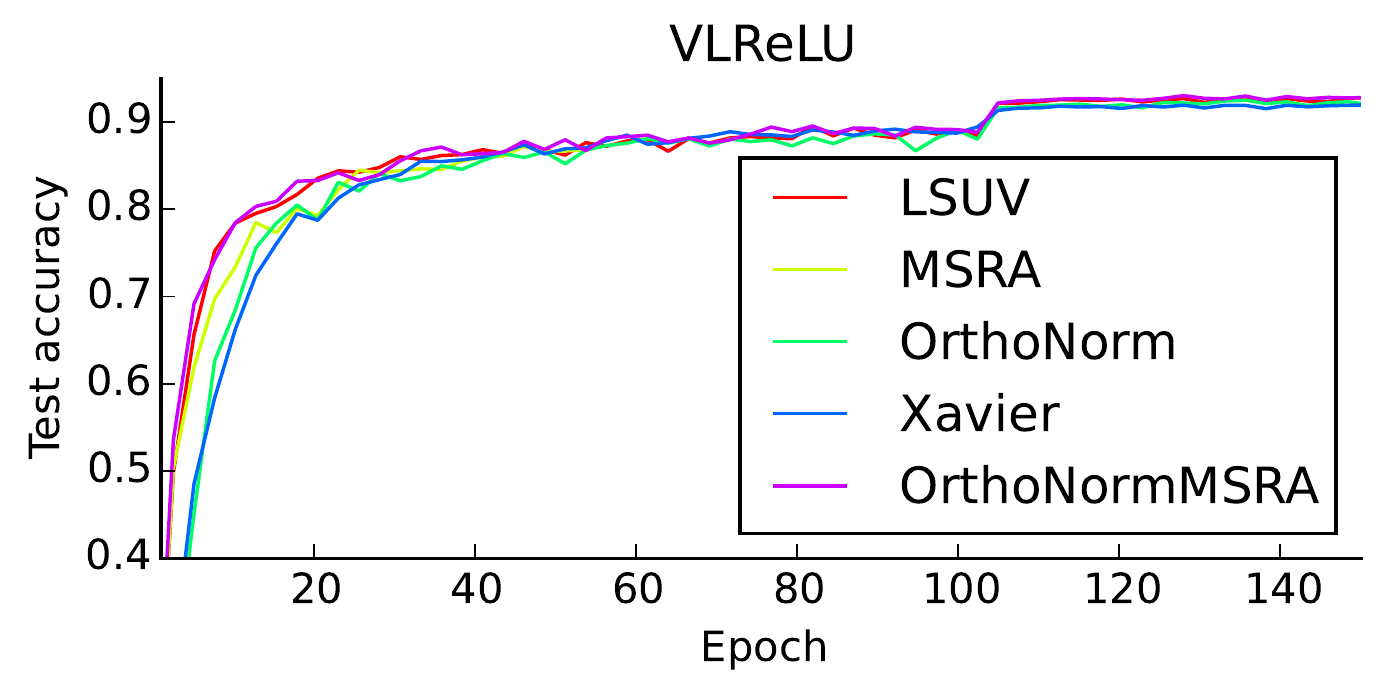}
\includegraphics[width=0.48\linewidth]{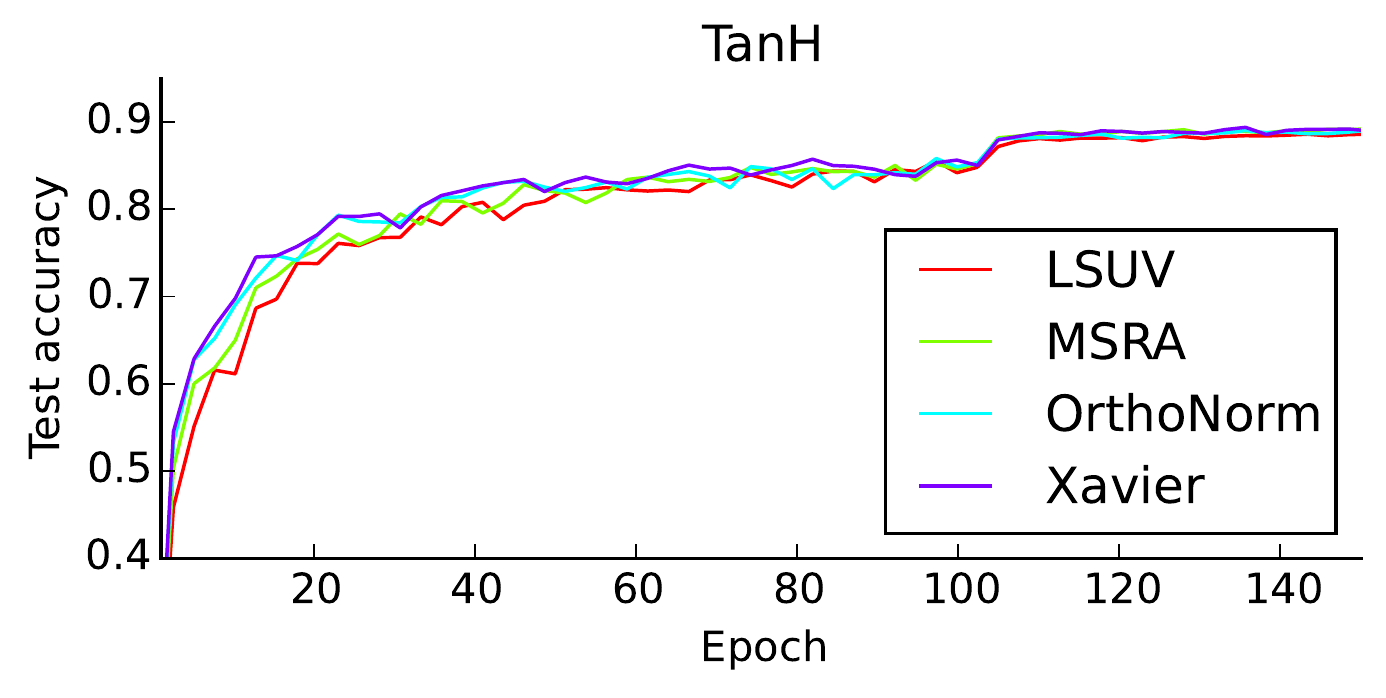}\\
\caption{CIFAR-10 accuracy of FitNet-4 with different activation functions. Note that graphs are cropped at 0.4 accuracy. Highway19 is the network from~\cite{Highway2015}.}
\label{fig:sgd-cifar10}
\end{figure}

Testing was performed on CIFAR-10 and results are in Table~\ref{tab:activations} and Figure~\ref{fig:sgd-cifar10}. Performance of orthonormal-based methods is superior to the scaled Gaussian-noise approaches for all tested types of activation functions, except tanh. Proposed LSUV strategy outperforms orthonormal initialization by smaller margin, but still consistently (see Table~\ref{tab:activations}). All the methods failed to train sigmoid-based very deep network. 
Figure~\ref{fig:sgd-cifar10} shows that LSUV method not only leads to better generalization error, but also converges faster for all tested activation functions, except tanh.

We have also tested how the different initializations work "out-of-the-box" with the  Residual net training~\cite{DeepResNet2015}; a residual net won the ILSVRC-2015 challenge. The original paper proposed different implementations of residual learning. We adopted the simplest one, showed in Table~\ref{tab:fitnet-architectures}, FitResNet-4. The output of each even convolutional layer is summed with the output of the previous non-linearity layer and then fed into the next non-linearity. Results are shown in Table~\ref{tab:activations-resnet}. LSUV is the only initialization algorithm which leads nets to convergence with all tested non-linearities without any additional tuning, except, again, sigmoid. 
It is worth nothing that the residual training improves results for ReLU and maxout, but does not help tanh-based network. 
\begin{table}[htb]
\caption{The performance of activation functions and initialization in the Residual learning setup~\cite{DeepResNet2015}, FitResNet-4 from Table~\ref{tab:fitnet-architectures}.The n/c symbol stands for ``failed to converge'';}
\label{tab:activations-resnet}
\centering
\begin{tabular}{llllll}
\hline
Init method & maxout & ReLU  & VLReLU & tanh & Sigmoid \\
\hline
LSUV           & \textbf{94.16} & \textbf{92.82} &  \textbf{93.36}& 89.17& n/c \\
OrthoNorm      & n/c  & 91.42    & n/c          & 89.31 & n/c \\
Xavier         & n/c  & 92.48    & \textbf{93.34} & \textbf{89.62} & n/c \\
MSRA           & n/c   & n/c     & n/c   & 88.59 & n/c \\
\hline
\end{tabular}
\end{table}

\subsection{Comparison to batch normalization (BN)}
LSUV procedure could be viewed as batch normalization of layer output done only before the start of training. Therefore, it is natural to compare LSUV against a batch-normalized network, initialized with the standard method.

\subsubsection{Where to put BN -- before or after non-linearity?}
It is not clear from the paper~\cite{BatchNorm2015} where to put the batch-normalization layer -- before input of each layer as stated in Section 3.1, or before non-linearity, as stated in section 3.2, so we have conducted an experiment with FitNet4 on CIFAR-10 to clarify this.

Results are shown in Table~\ref{tab:before-or-after}. Exact numbers vary from run to run, but in the most cases, batch normalization put after non-linearity performs better. 

\begin{table}[htb]
\centering
\caption{CIFAR-10 accuracy of batch-normalized FitNet4.\\ Comparison of batch normalization put before and after non-linearity.}
\label{tab:before-or-after}
\centering
\begin{tabular}{lrr}
\hline
Non-linearity & \multicolumn{2}{c}{Where to put BN}\\
& Before & After\\
\hline
TanH & 88.10 & 89.22\\
ReLU & 92.60 & 92.58\\
Maxout & 92.30 & 92.98 \\
\hline
\end{tabular}
\end{table}

\begin{figure}
\centering
\includegraphics[width=0.49\linewidth]{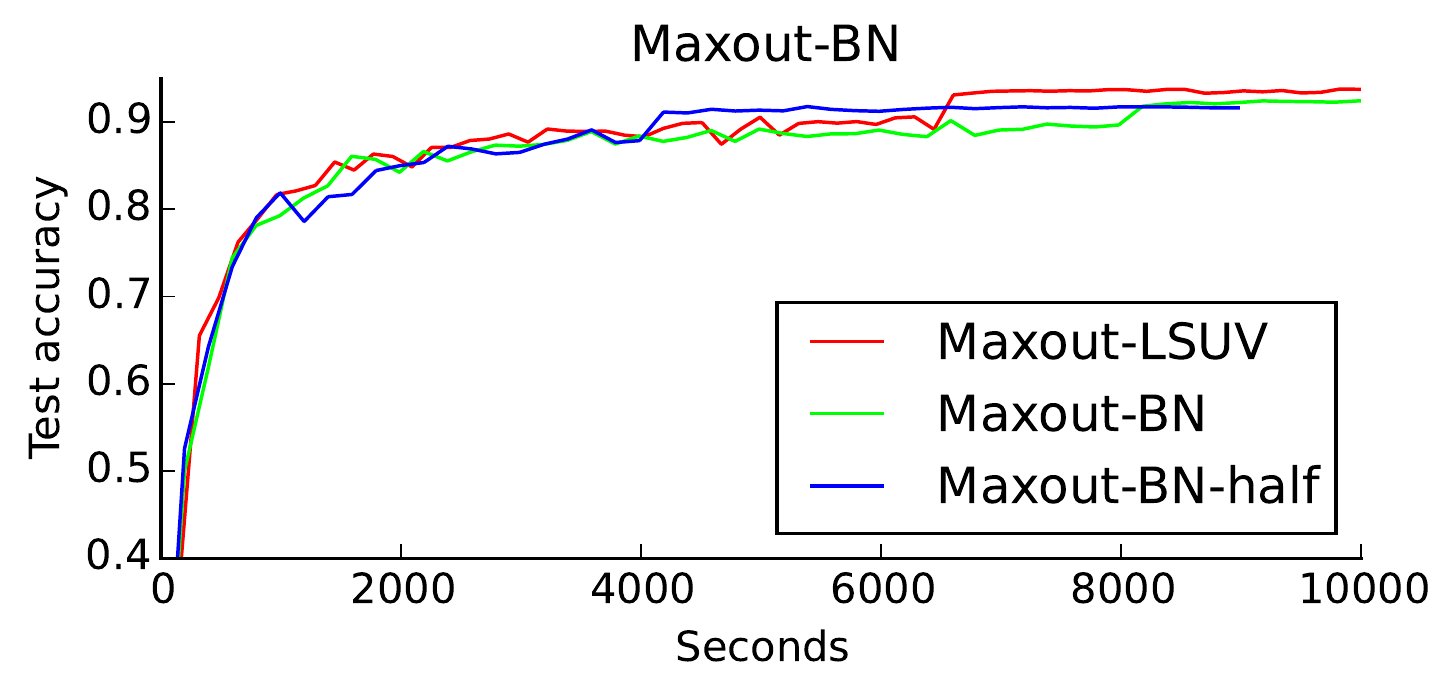}
\includegraphics[width=0.49\linewidth]{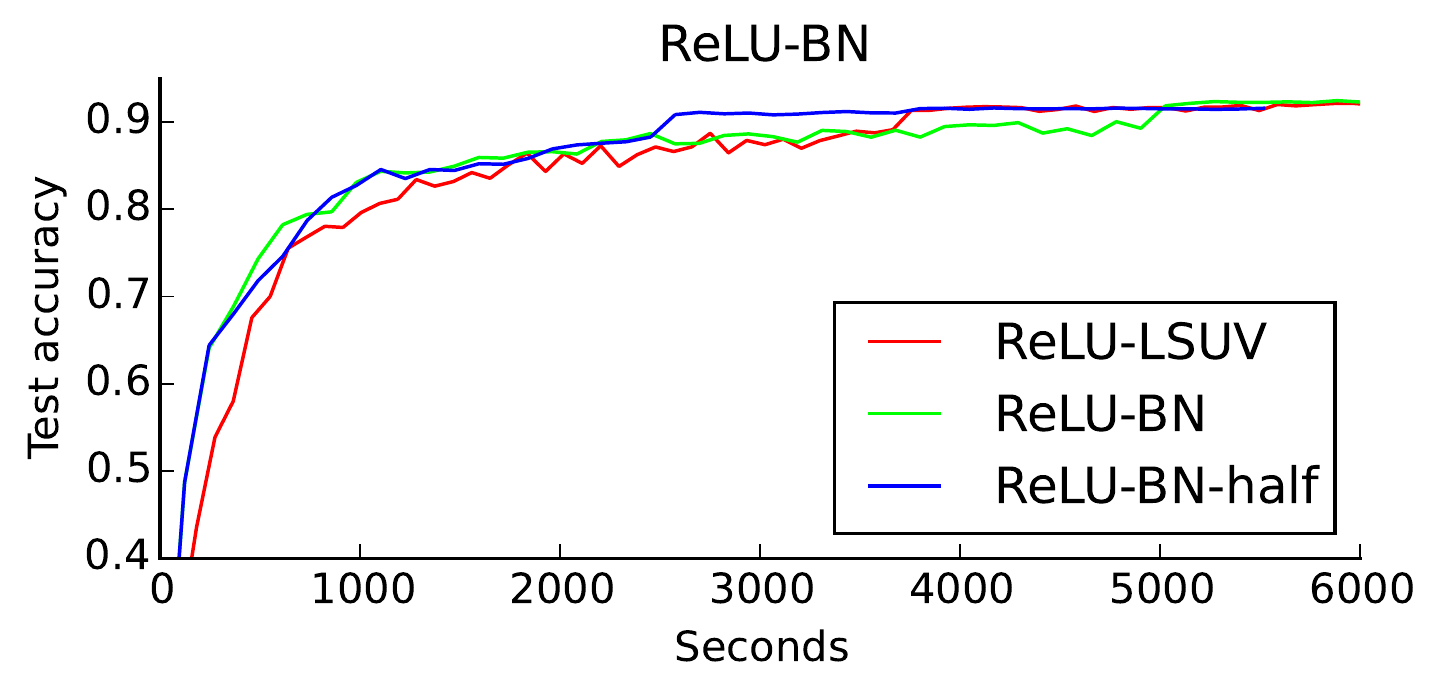}\\
\includegraphics[width=0.49\linewidth]{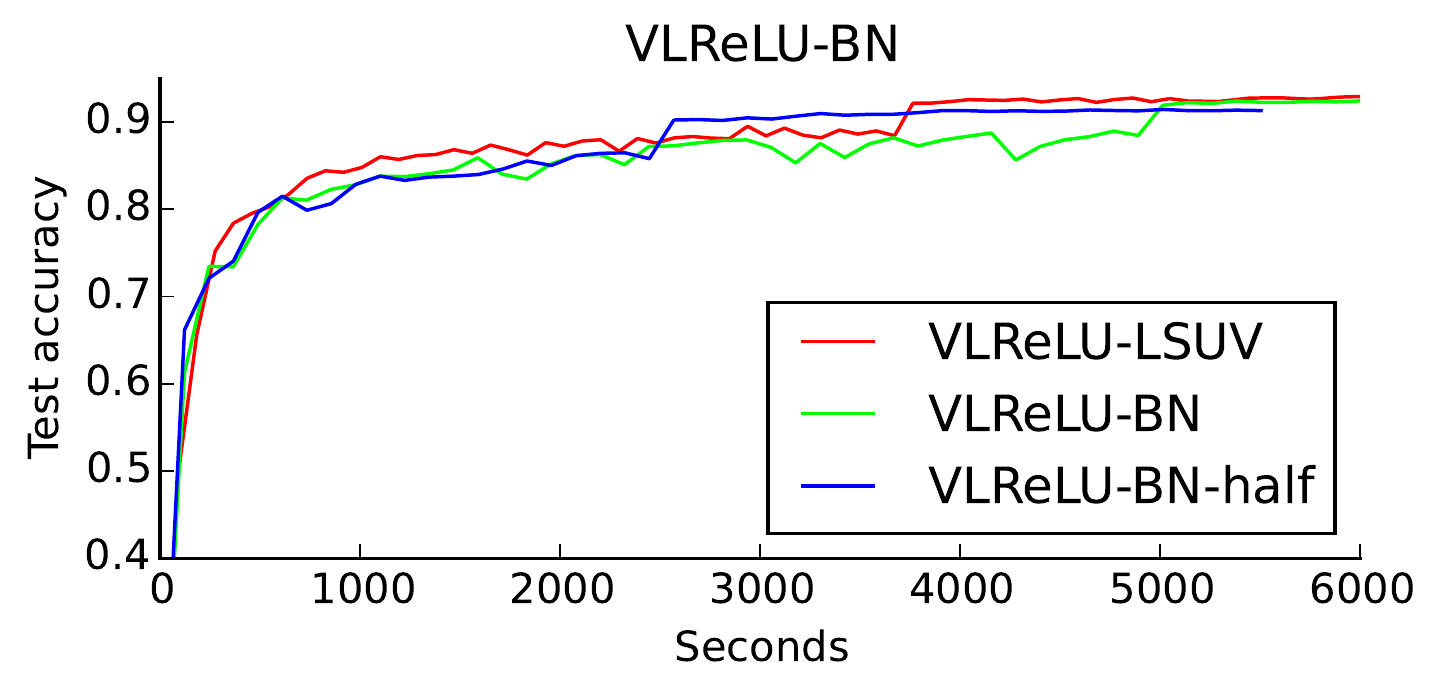}
\includegraphics[width=0.49\linewidth]{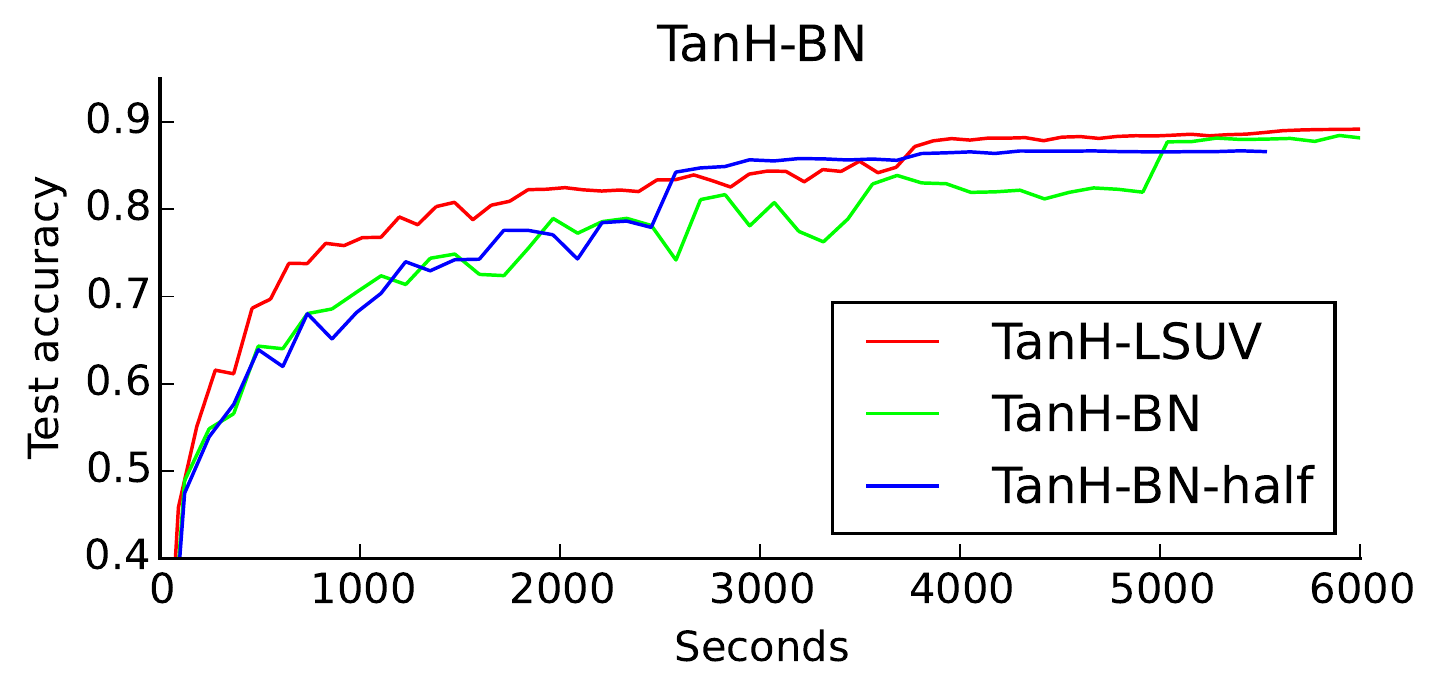}\\
\caption{CIFAR-10 accuracy of FitNet-4 LSUV and batch normalized~(BN) networks as function of wall-clock time. BN-half stands for half the number of iterations in each step.}
\label{fig:BN-vs-LSUV}
\end{figure}

In the next experiment we compare BN-FitNet4, initialized with Xavier and LSUV-initialized FitNet4. Batch-normalization reduces training time in terms of needed number of iterations, but each iteration becomes slower because of extra computations. 
The accuracy versus wall-clock-time graphs are shown in Figure~\ref{fig:BN-vs-LSUV}. LSUV-initialized network is as good as batch-normalized one. 

However, we are not claiming that batch normalization can always be replaced by proper initialization, especially in large datasets like ImageNet. 

\subsection{Imagenet training}
We trained CaffeNet (\cite{jia2014caffe}) and GoogLeNet~(\cite{Googlenet2015}) on the ImageNet-1000 dataset(~\cite{ILSVRC15}) with the original initialization and LSUV. CaffeNet is a variant of AlexNet with the nearly identical performance, where the order of pooling and normalization layers is switched to reduce the memory footprint. 

\begin{figure}
 \centering
\includegraphics[width=0.49\linewidth]{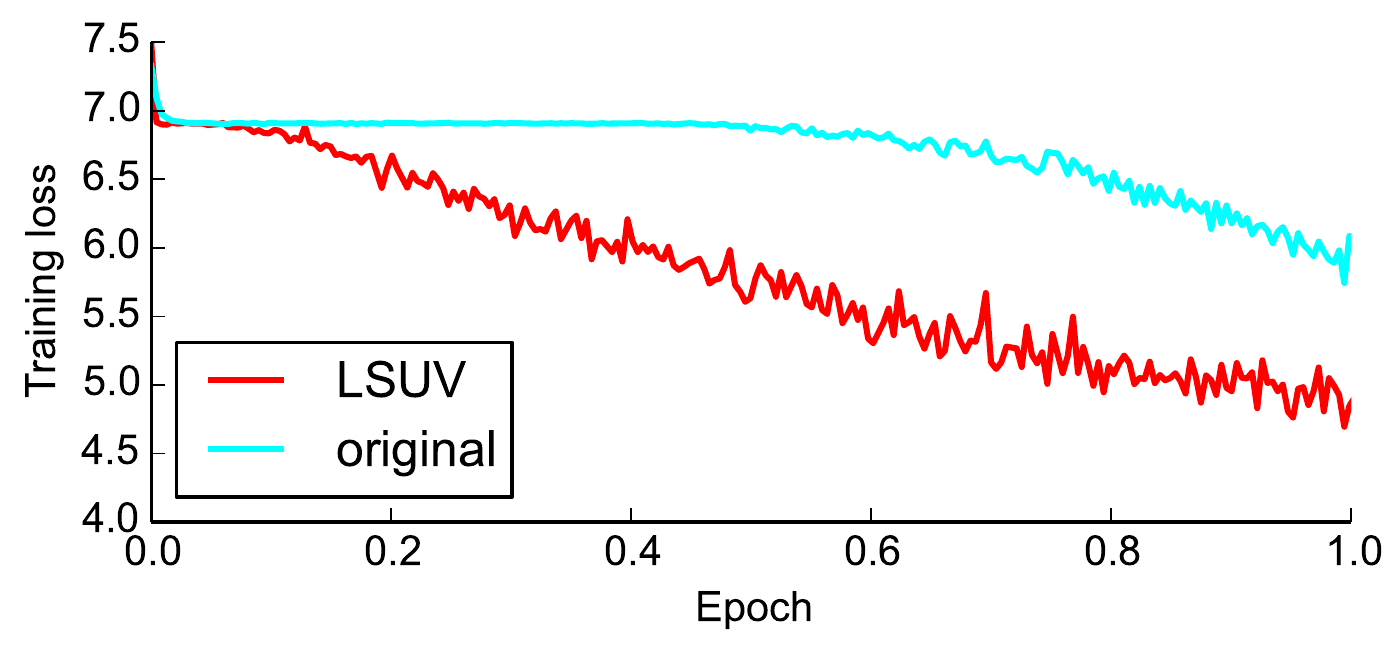}
\includegraphics[width=0.49\linewidth]{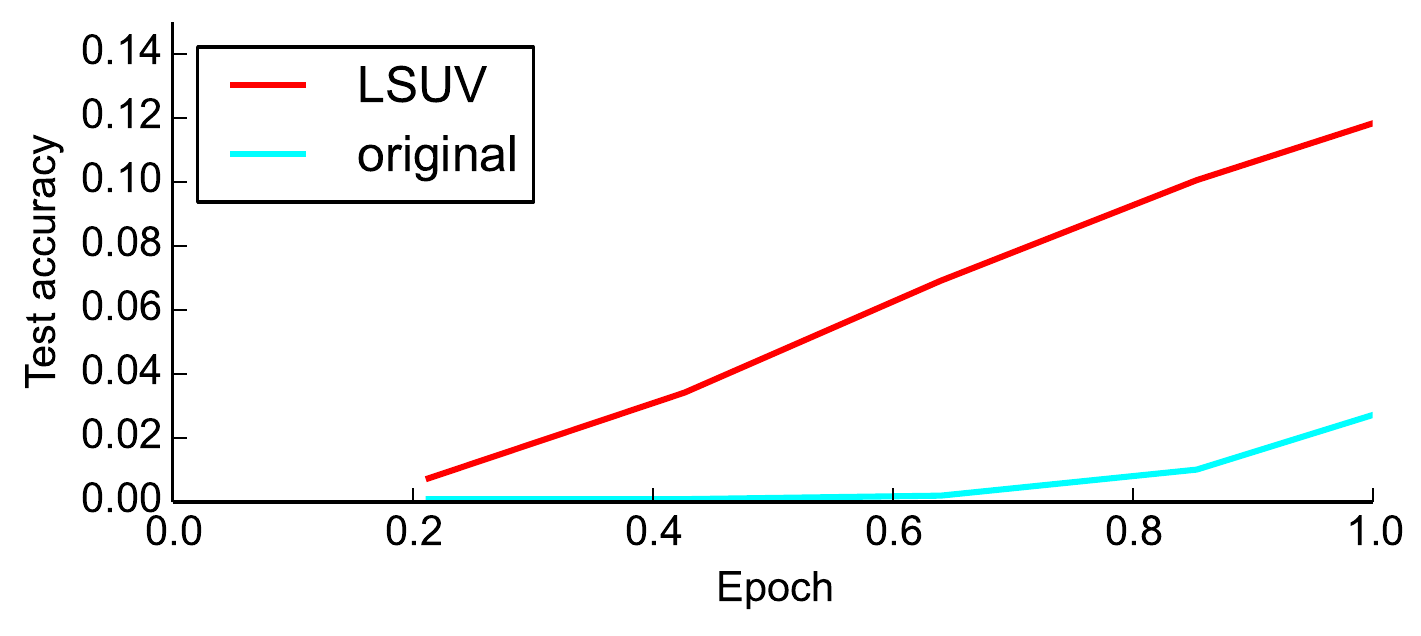}\\
\includegraphics[width=0.49\linewidth]{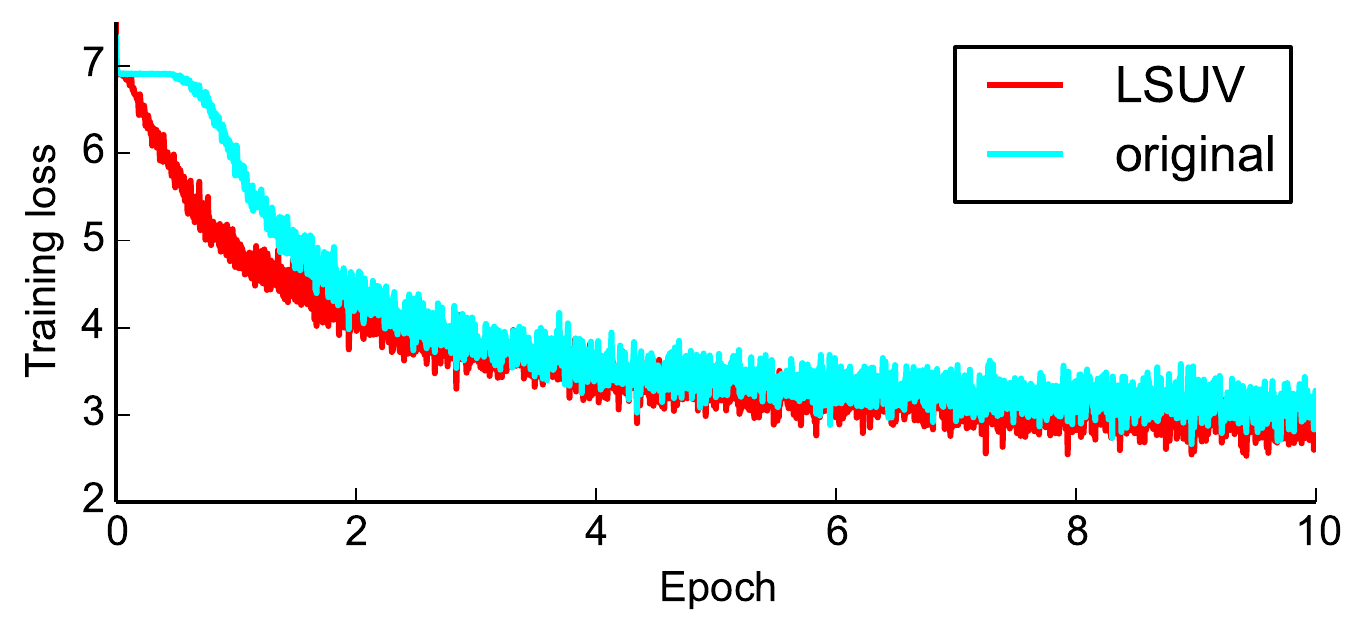}
\includegraphics[width=0.49\linewidth]{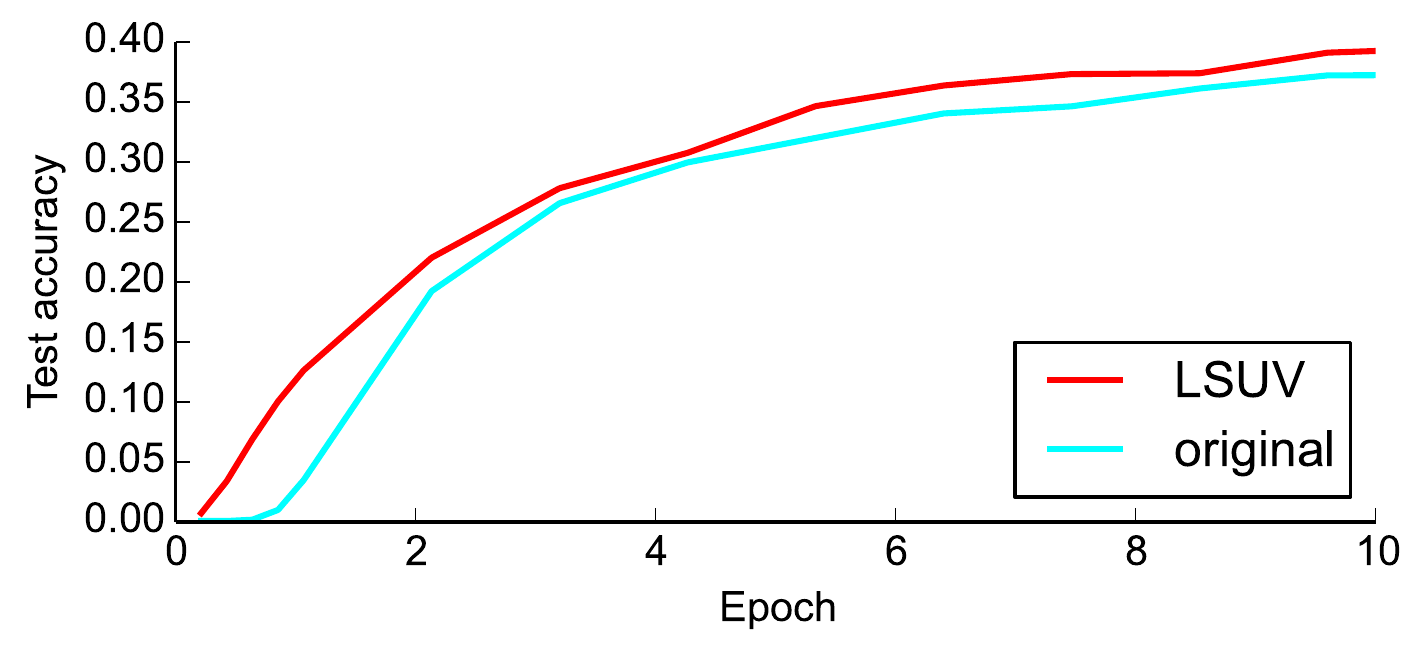}\\
\includegraphics[width=0.49\linewidth]{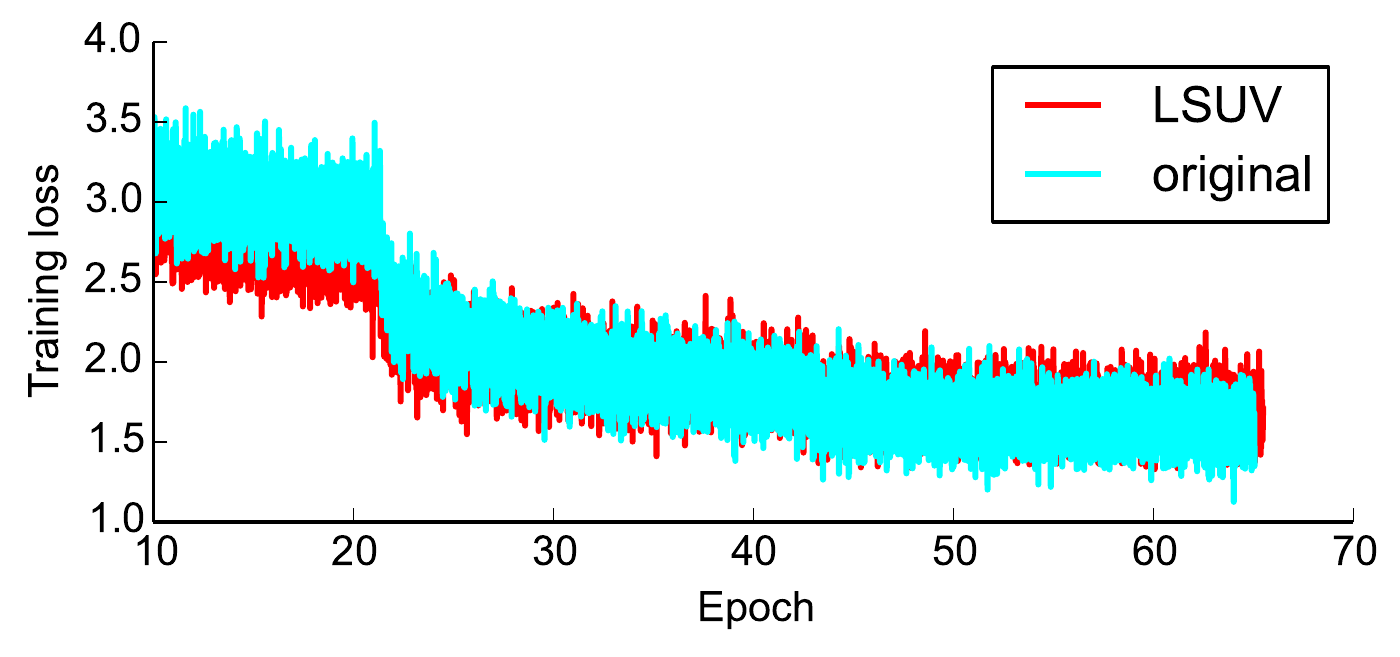}
\includegraphics[width=0.49\linewidth]{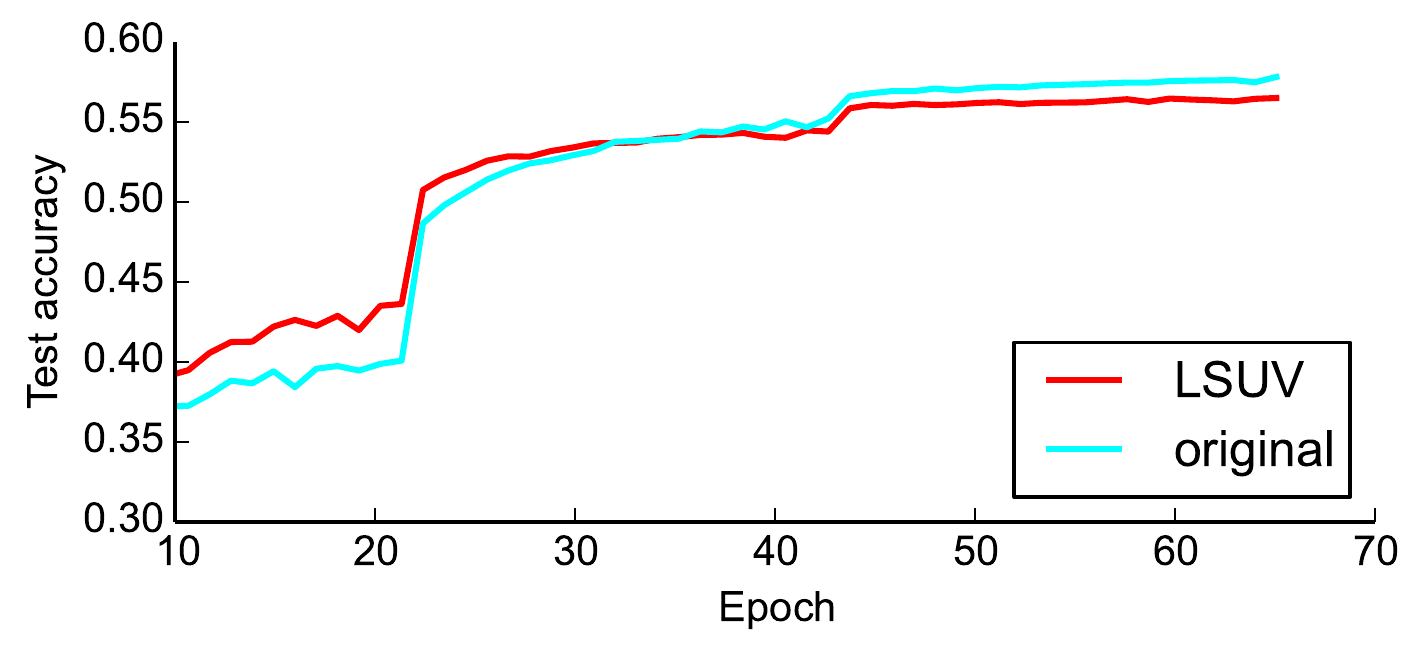}\\
\caption{CaffeNet training on ILSVRC-2012 dataset with LSUV and original~\cite{AlexNet2012} initialization. Training loss (left) and validation accuracy (right). Top -- first epoch, middle -- first 10 epochs, bottom -- full training.}
\label{fig:caffenet-training}
\end{figure}

LSUV initialization reduces the starting flat-loss time from 0.5 epochs to 0.05 for CaffeNet, and starts to converge faster, but it is overtaken by a standard CaffeNet at the 30-th epoch (see Figure~\ref{fig:caffenet-training}) and its final precision is 1.3\% lower. We have no explanation for this empirical phenomenon.

\begin{figure}
 \centering
\includegraphics[width=0.49\linewidth]{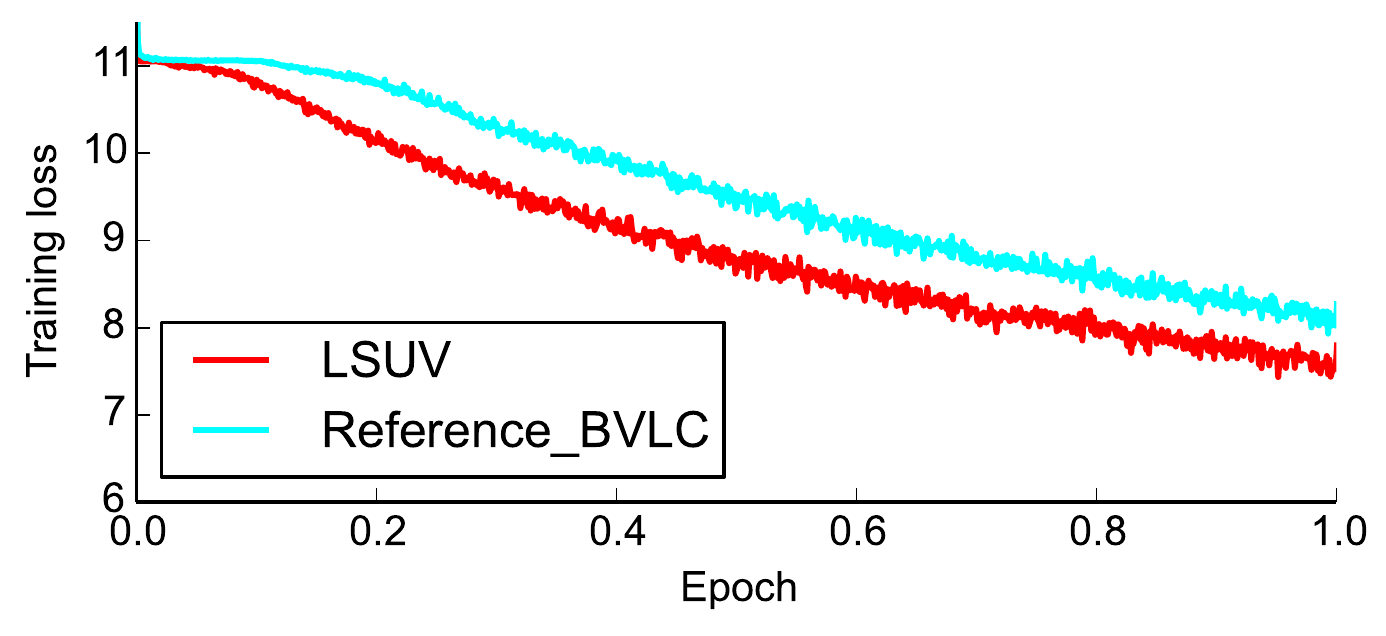}
\includegraphics[width=0.49\linewidth]{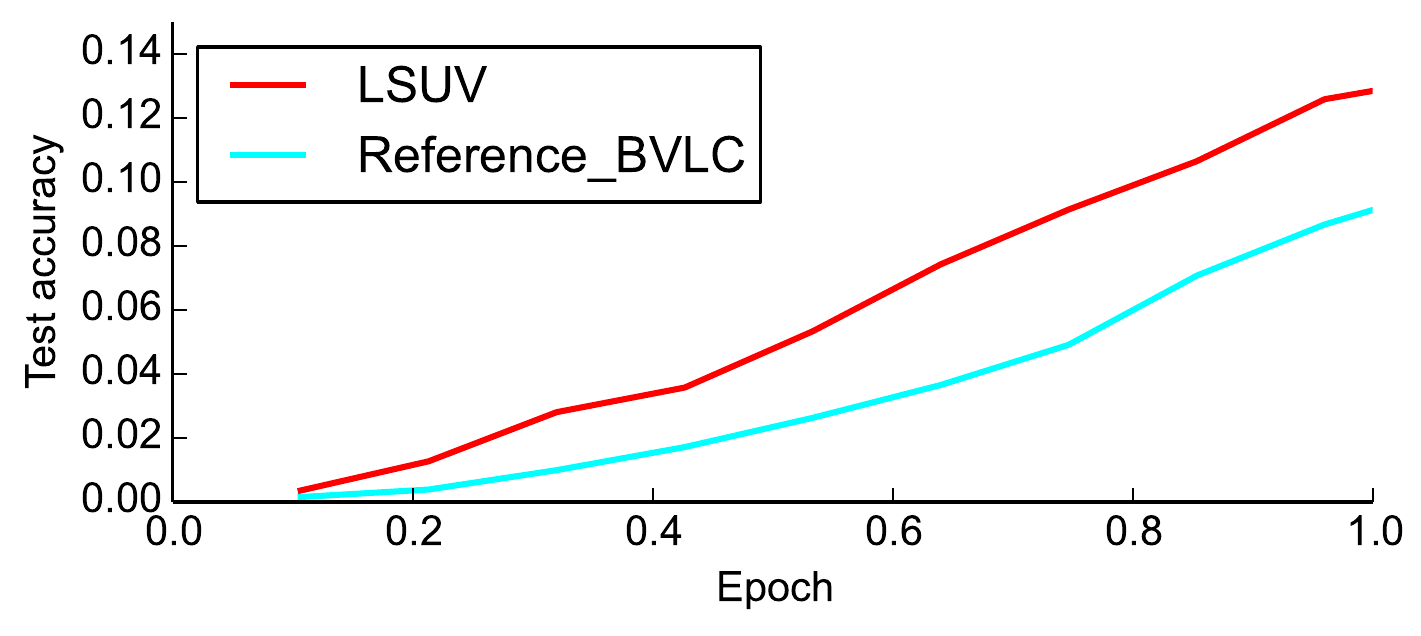}\\
\includegraphics[width=0.49\linewidth]{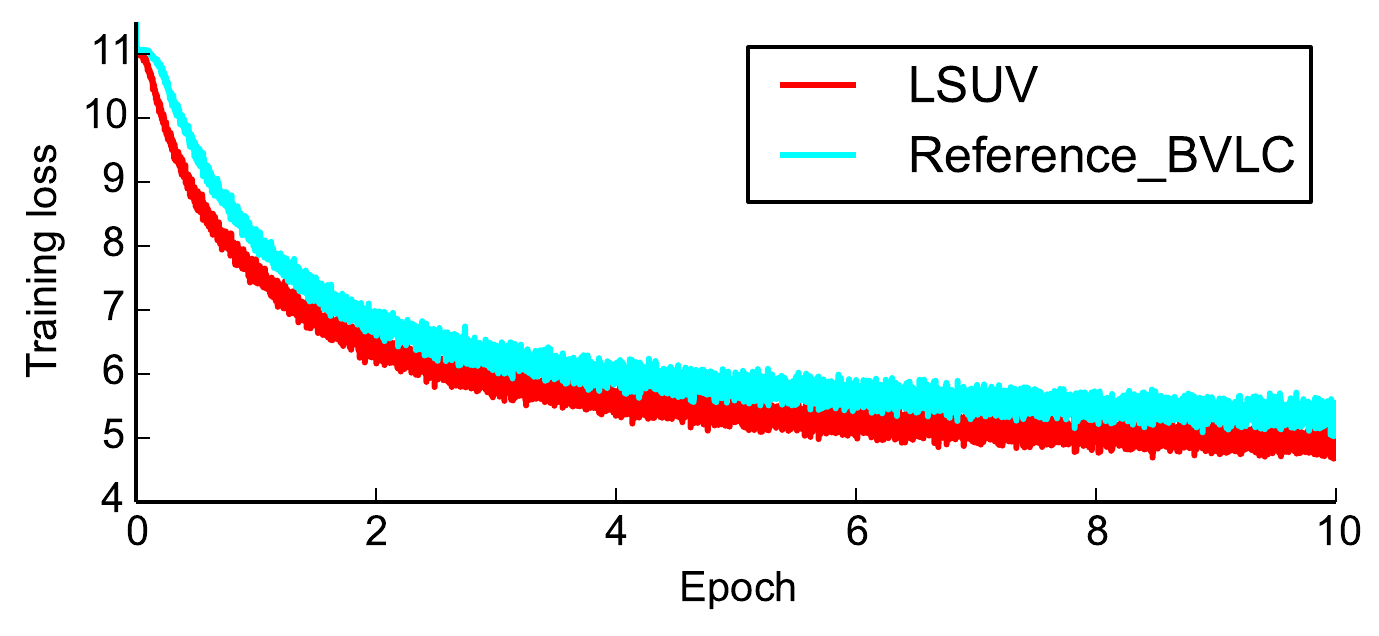}
\includegraphics[width=0.49\linewidth]{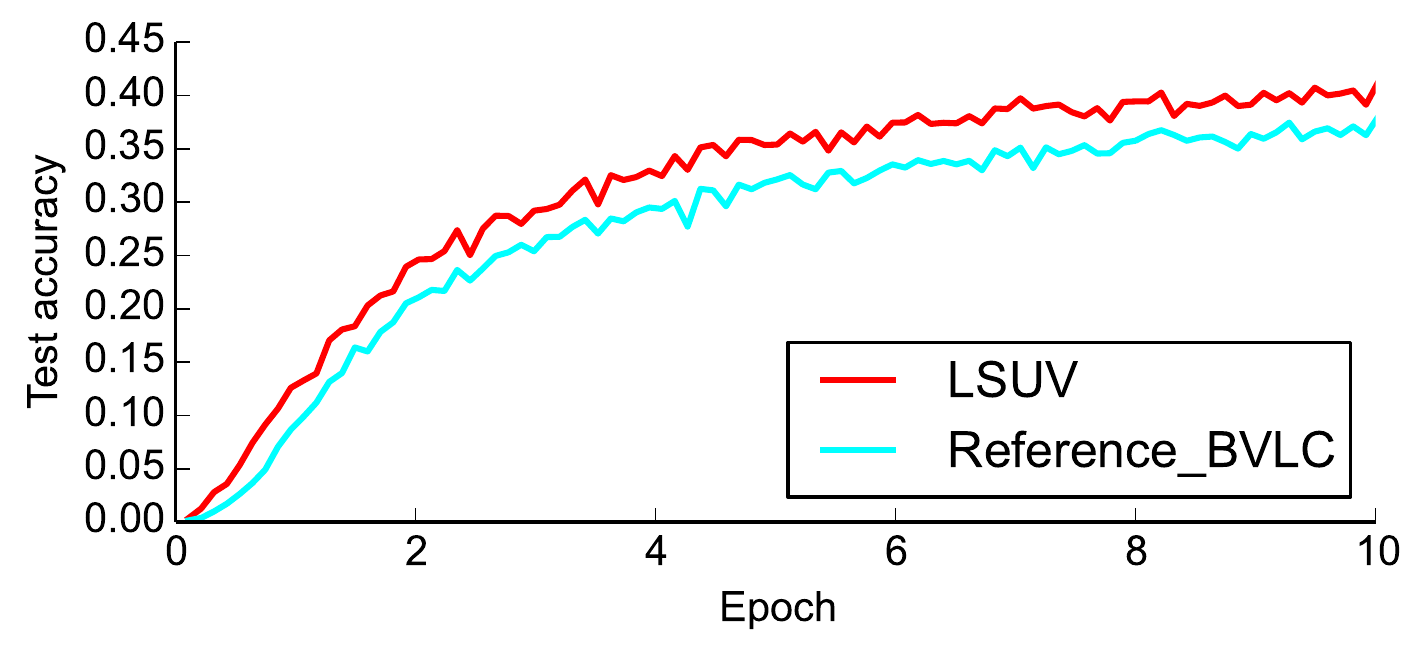}\\
\includegraphics[width=0.49\linewidth]{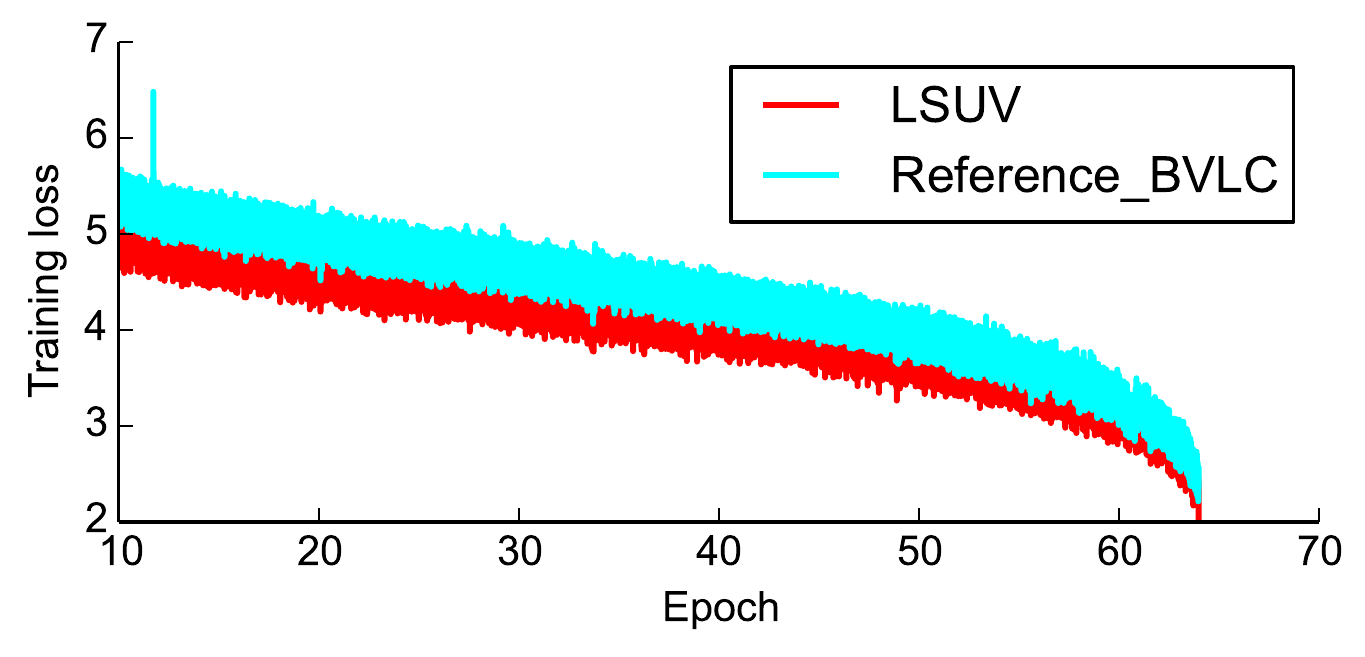}
\includegraphics[width=0.49\linewidth]{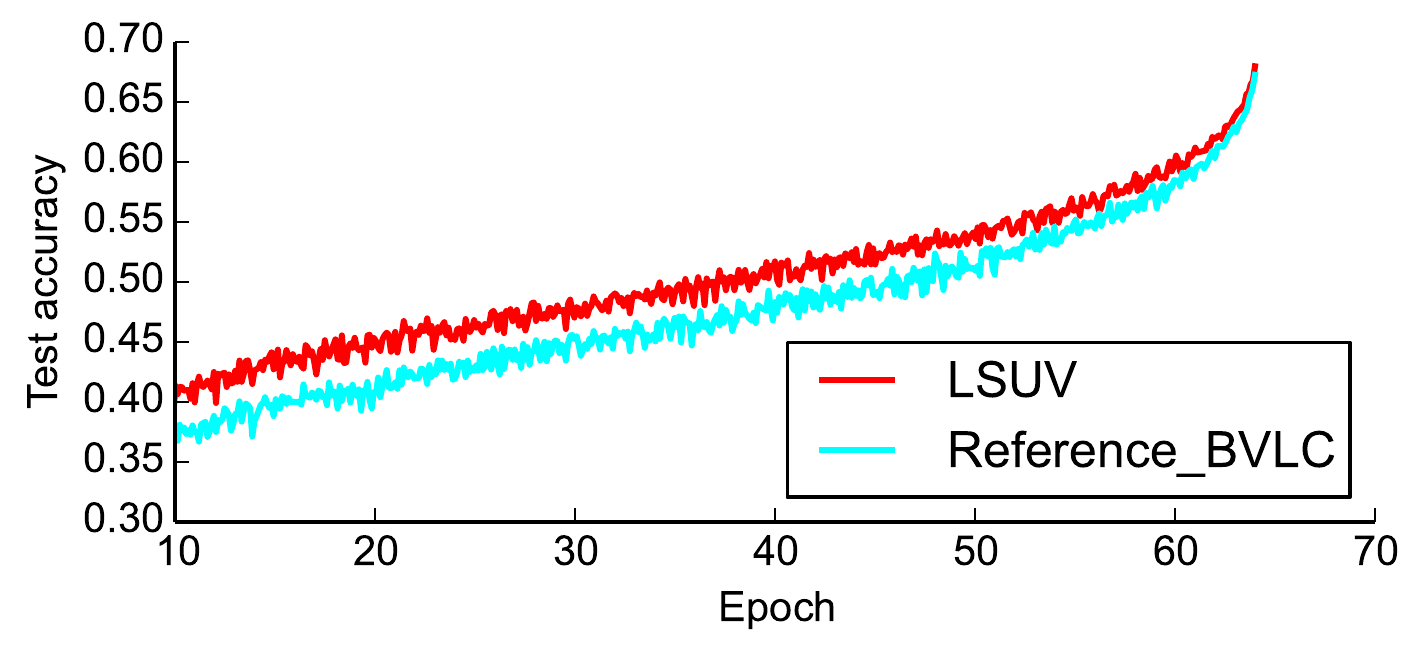}\\
\caption{GoogLeNet training on ILSVRC-2012 dataset with LSUV and reference~\cite{jia2014caffe} BVLC initializations. Training loss (left) and validation accuracy (right).  Top -- first epoch, middle -- first ten epochs, bottom -- full training}
\label{fig:googlenet-training}
\end{figure}

On the contrary, the LSUV-initialized GoogLeNet learns faster than hen then original one and shows better test accuracy all the time -- see Figure~\ref{fig:googlenet-training}. The final accuracy is 0.680 vs. 0.672 respectively.

\subsection{Timings}
\label{timings}
A significant part of LSUV initialization is SVD-decomposition of the weight matrices, e.g. for the fc6 layer of CaffeNet, an SVD of a 9216x4096 matrix is required. The computational overhead on top of generating almost instantly the scaled random Gaussian samples is shown in Table~\ref{tab:timings}. In the slowest case -- CaffeNet -- LSUV initialization takes 3.5 minutes, which is negligible in comparison the training time. 
\begin{table}[htb]
\caption{Time needed for network initialization \\ on top of random Gaussian (seconds).}
\label{tab:timings}
\centering
\begin{tabular}{r|rr|}
\hline
Network & \multicolumn{2}{c|}{Init}\\
& OrthoNorm & LSUV\\
\hline
FitNet4 & 1 & 4 \\
CaffeNet & 188 & 210 \\
GoogLeNet & 24 & 60 \\
\hline
\end{tabular}
\end{table}

\section{Conclusions}
\label{conclusions}
LSUV, layer sequential uniform variance, a simple strategy for weight initialization for deep net learning, is proposed.
We have showed that the LSUV initialization, described fully in six lines of pseudocode, is as good as complex learning schemes which need, for instance, auxiliary nets.

The LSUV initialization allows learning of very deep nets via standard SGD, is fast, and leads to (near) state-of-the-art results on MNIST, CIFAR, ImageNet datasets, outperforming the sophisticated systems designed specifically for very deep nets such as FitNets(~\cite{FitNets2014}) and Highway(~\cite{Highway2015}). The proposed initialization works well with different activation functions. 
  
 Our experiments confirm the finding of ~\cite{FitNets2014} that very thin, thus fast and low in parameters, but deep networks obtain comparable or even better performance than wider, but shallower ones. 
\subsubsection*{Acknowledgments}
The authors were supported by The Czech Science Foundation Project GACR P103/12/G084 and CTU student grant 
SGS15/155/OHK3/2T/13.
\bibliography{iclr2016_conference}
\bibliographystyle{iclr2016_conference}

\appendix
\section{Technical details}
\subsection{Influence of mini-batch size to LSUV initialization}
We have selected tanh activation as one, where LSUV initialization shows the worst performance and tested the influence of mini-batch size to training process. Note, that training mini-batch is the same for all initializations, the only difference is mini-batch used for variance estimation. One can see from Table~\ref{tab:batchsize} that there is no difference between small or large mini-batch, except extreme cases, where only two sample are used. 
\begin{table}[htb]
\caption{FitNet4 TanH final performance on CIFAR-10. Dependence on LSUV mini-batch size}
\label{tab:batchsize}
\centering
\begin{tabular}{l|lllll}
\hline
Batch size for LSUV&2 & 16 & 32 & 128 & 1024\\
Final accuracy, [\%] & 89.27 & 89.30 &89.30 &89.28 & 89.31 \\
\hline
\end{tabular}
\end{table}
\subsection{LSUV weight standard deviations in different networks}
Tables~\ref{tab:relu-std} and \ref{tab:lsuv-std} show the standard deviations of the filter weights, found by the LSUV procedure and by other initialization schemes. 
\begin{table}[htb]
\caption{Standard deviations of the weights per layer for different initializations, FitNet4, CIFAR10, ReLU}
\label{tab:relu-std}
\centering
\begin{tabular}{lllll}
\hline
Layer & LSUV & OrthoNorm & MSRA & Xavier\\
\hline
conv11 & 0.383 & 0.175 & 0.265  & 0.191 \\
conv12 & 0.091 & 0.058 & 0.082   & 0.059\\
conv13 & 0.083 & 0.058 & 0.083 & 0.059\\
conv14 & 0.076  & 0.058 & 0.083  & 0.059\\
conv15 & 0.068  & 0.048 & 0.060  & 0.048\\
\hline
conv21 & 0.036  & 0.048 & 0.052  & 0.037\\
conv22 & 0.048  & 0.037 & 0.052  & 0.037\\
conv23 & 0.061  & 0.037& 0.052  & 0.037\\
conv24 & 0.052  & 0.037& 0.052 & 0.037\\
conv25 & 0.067 & 0.037& 0.052  & 0.037\\
conv26 & 0.055 & 0.037& 0.052  & 0.037\\
\hline
conv31 & 0.034  &0.037& 0.052  & 0.037\\
conv32 & 0.044  &0.029& 0.041& 0.029\\
conv33 & 0.042  &0.029& 0.041 & 0.029\\
conv34 & 0.041 &0.029& 0.041  & 0.029\\
conv35 & 0.040  &0.029& 0.041 & 0.029\\
conv36 & 0.043 &0.029& 0.041 & 0.029\\
\hline
ip1 & 0.048 & 0.044 & 0.124  &  0.088\\
\hline
\end{tabular}
\end{table}
\begin{table}[htb]
\caption{Standard deviations of the weights per layer for different non-linearities, found by LSUV, FitNet4, CIFAR10}
\label{tab:lsuv-std}
\centering
\begin{tabular}{lllll}
\hline
Layer&TanH&ReLU&VLReLU&Maxout\\
\hline
conv11&0.386&0.388&0.384&0.383\\
conv12&0.118&0.083&0.084&0.058\\
conv13&0.102&0.096&0.075&0.063\\
conv14&0.101&0.082&0.080&0.065\\
conv15&0.081&0.064&0.065&0.044\\
\hline
conv21&0.065&0.044&0.037&0.034\\
conv22&0.064&0.055&0.047&0.040\\
conv23&0.060&0.055&0.049&0.032\\
conv24&0.058&0.064&0.049&0.041\\
conv25&0.061&0.061&0.043&0.040\\
conv26&0.063&0.049&0.052&0.037\\
\hline
conv31&0.054&0.032&0.037&0.027\\
conv32&0.052&0.049&0.037&0.031\\
conv33&0.051&0.048&0.042&0.033\\
conv34&0.050&0.047&0.038&0.028\\
conv35&0.051&0.047&0.039&0.030\\
conv36&0.051&0.040&0.037&0.033\\
\hline
ip1&0.084&0.044&0.044&0.038\\
\hline
\end{tabular}
\end{table}
\subsection{Gradients}
To check how the activation variance normalization influences the variance of the gradient, we measure the average variance of the gradient at all layers after 10 mini-batches. The variance is close to $10^{-9}$ for all convolutional layers. It is much more stable than for the reference methods, except MSRA; see Table~\ref{tab:grad-var}. 
\begin{table}[htb]
\caption{Variance of the initial gradients per layer, different initializations, FitNet4, ReLU}
\label{tab:grad-var}
\centering
\begin{tabular}{lllll}
\hline
Layer&LSUV&MSRA&OrthoInit&Xavier\\
\hline
conv11&4.87E-10&9.42E-09&5.67E-15&2.30E-14\\
conv12&5.07E-10&9.62E-09&1.17E-14&4.85E-14\\
conv13&4.36E-10&1.07E-08&2.30E-14&9.94E-14\\
conv14&3.21E-10&7.03E-09&2.95E-14&1.35E-13\\
conv15&3.85E-10&6.57E-09&6.71E-14&3.10E-13\\
\hline
conv21&1.25E-09&9.11E-09&1.95E-13&8.00E-13\\
conv22&1.15E-09&9.73E-09&3.79E-13&1.56E-12\\
conv23&1.19E-09&1.07E-08&8.18E-13&3.28E-12\\
conv24&9.12E-10&1.07E-08&1.79E-12&6.69E-12\\
conv25&7.45E-10&1.09E-08&4.04E-12&1.36E-11\\
conv26&8.21E-10&1.15E-08&8.36E-12&2.99E-11\\
\hline
conv31&3.06E-09&1.92E-08&2.65E-11&1.05E-10\\
conv32&2.57E-09&2.01E-08&5.95E-11&2.28E-10\\
conv33&2.40E-09&1.99E-08&1.21E-10&4.69E-10\\
conv34&2.19E-09&2.25E-08&2.64E-10&1.01E-09\\
conv35&1.94E-09&2.57E-08&5.89E-10&2.27E-09\\
conv36&2.31E-09&2.97E-08&1.32E-09&5.57E-09\\
\hline
ip1&1.24E-07&1.95E-07&6.91E-08&7.31E-08\\
\hline
var(ip1)/var(conv11)&255&20&12198922&3176821\\
\hline
\end{tabular}
\end{table}
\end{document}